\documentclass[11pt]{article}

\usepackage[final]{acl}

\usepackage{times}
\usepackage{latexsym}
\usepackage{subfigure}
\usepackage{multirow}

\usepackage{booktabs}
\usepackage{amsmath}
\usepackage{colortbl}

\usepackage{tcolorbox}
\newtcolorbox{examplebox}{
  colback=gray!10,
  colframe=gray!50,
  arc=4pt,
  boxrule=0.4pt,
  left=6pt,
  right=6pt,
  top=4pt,
  bottom=4pt,
  fontupper=\scriptsize\ttfamily  
} % \scriptsize \footnotesize \small  \normalsize  \sffamily  \ttfamily

\usepackage[T1]{fontenc}
\usepackage[utf8]{inputenc}

\usepackage{microtype}

\usepackage{inconsolata}

\usepackage{graphicx}

\usepackage{cleveref}

\usepackage{enumitem}

\title{Fusion Training for Mathematical Generalization in\\ Large Language Models}

\author{
 \textbf{Congfeng Cao\textsuperscript{1}},
 \textbf{Pengyu Zhang\textsuperscript{2}},
 \textbf{Jelke Bloem\textsuperscript{1}}
\\
\\
 \textsuperscript{1}Institute for Logic, Language and Computation, University of Amsterdam
 \\
 \textsuperscript{2}INDE Lab, University of Amsterdam,
}

\begin{document}
\maketitle

\begin{abstract}

Thinking Mode Fusion (TMF) enables large language models to support both concise responses and long-form reasoning by unifying a non-thinking mode and a thinking mode within a single model. However, its training dynamics, including the \emph{data ratio} and \emph{training schedule} between the two modes, remain underexplored.
In this work, we present a systematic study of TMF by analyzing the effects of the training schedule and data ratio between thinking and non-thinking modes. 
Focusing on mathematical problem solving, we construct a benchmark with multiple thinking-to-non-thinking data ratios and three training schedules. 
Our results reveal an asymmetric interaction between the two modes: increasing the ratio of non-thinking supervision reduces the accuracy of the thinking mode.
We further show that different training schedules modulate this trade-off and that the optimal schedule depends on the data ratio. 
Finally, we quantify a negative correlation between non-thinking and thinking mode supervision, highlighting an inherent tension between these two modes. 
These findings provide practical guidance for designing effective TMF training settings.
All code and data are released to support further research at: \href{https://github.com/caocongfeng/Fusion-Bench.git}{\textbf{Fusion Bench}}.
\end{abstract}

\section{Introduction}\label{Sec:Introduction}

Large language models (LLMs) have made substantial progress in \emph{long-form reasoning} and have been increasingly applied to complex tasks via extended chains of thought~\citep{NEURIPS2022_9d560961,NEURIPS2022_8bb0d291,wang2023selfconsistency}.
However, many real-world queries are comparatively simple and do not require long-form reasoning.
For such cases, long-form reasoning can be unnecessarily time-consuming and computationally expensive.

\begin{figure}[t]
\centering
\includegraphics[width=0.75\linewidth]{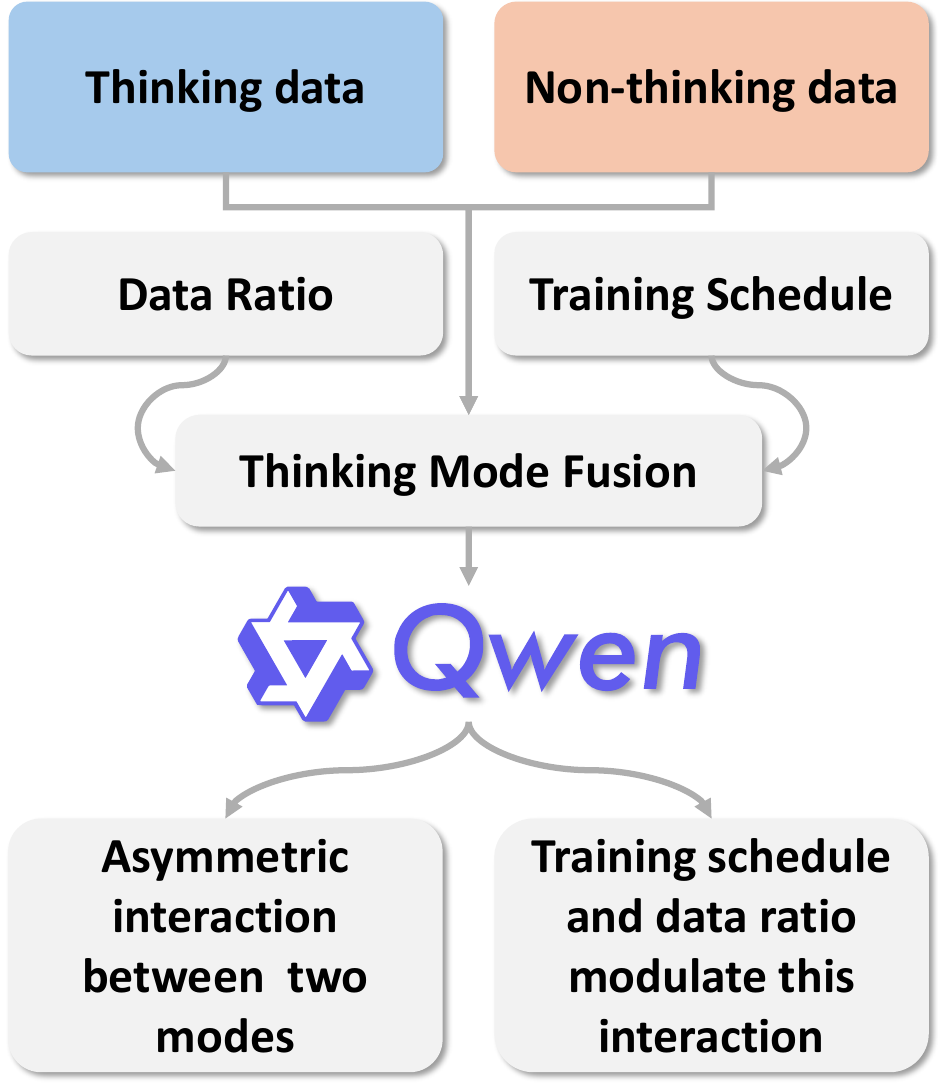}
\caption{An overview of the experimental design for analyzing Thinking Mode Fusion under different training orders and data ratios.}
\label{fig:fig1}
\end{figure}

% $revision$
As a result, existing efforts have explored switching models between simple queries and complex problems~\citep{openai2024openaio1card,openai2025openaio1card}, or adapting the length of model responses to improve reasoning efficiency~\citep{arora2025training, team2025kimi, shen-etal-2025-dast, wu2025unlocking, zhang-etal-2025-adaptthink}.
Qwen3 series models address this limitation by introducing \textbf{Thinking Mode Fusion (TMF)} during post-training~\citep{yang2025qwen3technicalreport}.

% As a result, practical systems often switch between chat-optimized models for simple queries and reasoning-oriented models for complex, multi-step problems~\citep{openai2024openaio1card,openai2025openaio1card,yang2025qwen3technicalreport}.
% AdaptThink proposes teaching reasoning models to choose the optimal thinking mode adaptively based on problem difficulty~\citep{zhang-etal-2025-adaptthink}.
% Qwen3 series models~\footnote{https://qwen.ai} address this limitation by introducing \textbf{Thinking Mode Fusion (TMF)} during post-training~\citep{yang2025qwen3technicalreport}.
% TMF unifies a \emph{thinking} mode (long multi-step reasoning) and a \emph{non-thinking} mode (fast, concise answering) within a single model, enabling users to control reasoning behavior without deploying separate systems.
% revise
TMF unifies a \emph{thinking} mode and a \emph{non-thinking} mode within a single model, enabling users to control reasoning behavior for both long-form reasoning and concise responses.
% revise here to further explain TMF
TMF is implemented through two different chat templates, i.e., a thinking chat template and a non-thinking chat template, that support mode selection.
% add
% In practice, TMF is implemented through continual supervised fine-tuning combined with a unified chat template that supports mode selection.
% 
This design reduces deployment complexity and inference costs while allowing the model to adjust its reasoning depth under different computational budgets.

Despite Qwen3's strong empirical performance, the mechanisms of TMF and the interaction between the two modes remain underexplored.
In particular, it remains unclear how the \emph{training schedule}, i.e., the training order of the two modes, and the \emph{data ratio}, i.e., the combination of thinking and non-thinking data, affect overall model performance.
% In particular, the two modes hold different and potentially competing objectives: the non-thinking mode favors concise responses, whereas the thinking mode emphasizes long-form reasoning responses.
This raises a practical question: \textbf{when both modes are trained within a single model, how do the training schedule and the data ratio affect the model's performance?}

To investigate this question, we focus on mathematics, where problem difficulty varies widely, from short-solution questions to Olympiad-level tasks that require long-form reasoning. 
As shown in~\Cref{fig:fig1}, we systematically investigate the interaction between the two modes and analyze the effects of TMF under different training schedules and varying ratios of thinking to non-thinking data. 
To support this study, we construct a benchmark, \textbf{Fusion Bench}, in accordance with the Qwen3 design and consisting of two types of mathematical problems: simple problems for non-thinking mode training and Olympiad-level problems for thinking mode training under diverse data ratios and training mode combinations.
Our results reveal an asymmetric interaction between the two modes: 
% the trade-off between the two modes shows that thinking mode training results in performance degradation as non-thinking supervision increases.
the trade-off between the two modes shows that thinking mode performance degrades as non-thinking supervision increases.
% add
Note that TMF unifies two modes (long-form reasoning and concise response) within a single model for a single task, which is different from multi-task settings such as math and code.
% Note TMF is template-controled methods which is different from multi-task, such as math and code. 
Moreover, we show that the training schedule modulates this trade-off and that the optimal schedule depends on the data ratio.
Finally, we quantify a negative correlation between non-thinking and thinking mode supervision, highlighting an inherent tension between these two modes. 

Our main contributions are as follows:

\begin{itemize}
    \item We conduct a systematic study of training schedules and data ratios in TMF, revealing a trade-off between thinking and non-thinking modes, where increased non-thinking supervision degrades thinking performance, and we quantify a negative correlation between the two modes.
    \item We show that training schedules modulate this trade-off: while the optimal schedule depends on the data ratio, the Mix training schedule consistently achieves strong performance across both thinking and non-thinking modes when averaged over different ratios.
    \item We construct and release Fusion Bench~\footnote{\href{https://anonymous.4open.science/r/Fusion-Bench-2D62}{https://anonymous.4open.science/r/Fusion-Bench-2D62}}, a benchmark comprising thinking and non-thinking datasets with varying data ratios and training schedules, enabling controlled and reproducible analysis of TMF.

\end{itemize}

% \begin{itemize}
%     \item We present a systematic study of the effects of the training schedule and the data ratio in TMF, revealing a trade-off between the two modes, in which increased non-thinking supervision degrades thinking mode performance.
%     \item We construct a benchmark, \textbf{Fusion Bench}, containing non-thinking and thinking datasets with varying data ratios and training combinations to facilitate controlled TMF analysis. We release our benchmark and code to the research community.\footnote{\href{https://anonymous.4open.science/r/Fusion-Bench-2D62}{https://anonymous.4open.science/r/Fusion-Bench-2D62}}
% \end{itemize}
% We present a systematic study of the effects of the training schedule and the data ratio in TMF, revealing a trade-off between the two modes, in which increased non-thinking supervision degrades thinking mode performance.
% We further construct a benchmark, \textbf{Fusion Bench}, containing paired non-thinking and thinking datasets with varying data ratios and training combinations to facilitate controlled TMF analysis.
% Finally, we release our benchmark and code to the research community.
% \footnote{\href{https://anonymous.4open.science/r/Fusion-Bench-2D62}{https://anonymous.4open.science/r/Fusion-Bench-2D62}}

\section{Related Work}\label{Sec:Related Work}
% \paragraph{Qwen3 Model.}
% \paragraph{Fusion Thinking and Data Mixing Training.}

\paragraph{Thinking Mode Fusion and Hybrid Thinking.}
Large language models (LLMs) have shown that supervised fine-tuning (SFT) on long chain-of-thought~\citep{NEURIPS2022_9d560961} traces is able to improve performance on complex tasks.

However, enhancing reasoning depth through specialized post-training increases computational costs due to the long reasoning responses. Models such as OpenAI o1~\citep{openai2024openaio1card}, DeepSeek-R1~\citep{deepseekai2025deepseekr1incentivizingreasoningcapability}, and Gemini~\citep{gemmateam2025gemma3technicalreport} widely adopt hybrid thinking mechanisms that control whether the model engages in reasoning, thereby achieving a more efficient and flexible reasoning process. 
Existing efforts to improve reasoning efficiency focus on reducing the length of model responses~\citep{arora2025training, team2025kimi, shen-etal-2025-dast}.
\citet{wu2025unlocking} merge reasoning and non-reasoning models to reduce output length.
AdaptThink proposes teaching reasoning models to choose the optimal thinking mode adaptively based on problem difficulty~\citep{zhang-etal-2025-adaptthink}.
Qwen3~\citep{yang2025qwen3technicalreport} introduced Thinking Mode Fusion (TMF), which allows a single model to support both long-form reasoning and concise responses controlled by the chat format.
\citet{wang2025demystifyinghybridthinkingllms} further demystify hybrid thinking, revealing that current hybrid thinking LLMs only achieve partial mode separation, where reasoning behaviors often leak into the no-think mode. However, the mechanisms of TMF remain underexplored.

% Large language models enhance reasoning depth through specialized post-training but increase the computational expense for long reasoning. Models such as OpenAI o1~\citep{openai2024openaio1card}, DeepSeek-R1~\citep{deepseekai2025deepseekr1incentivizingreasoningcapability}, and
% Gemini~\citep{gemmateam2025gemma3technicalreport} widely adopt hybrid thinking that controls whether the model engages in reasoning, thereby achieving a more efficient and flexible reasoning process. Building on this, Qwen3 \citep{yang2025qwen3technicalreport} introduced Thinking Mode Fusion (TMF), which allows a single model to support both long-form reasoning and concise responses. \citet{wang2025demystifyinghybridthinkingllms} demystify hybrid thinking, revealing that current hybrid thinking LLMs only achieve partial mode separation: reasoning behaviors often leak into the no-think mode.

\paragraph{Data Mixture.}
Prior studies have shown that data mixture plays a critical role in both pre-training and post-training stages.
\citet{ye2025data} outlined the quantitative predictability of model performance with respect to mixture proportions. \citet{liu2025regmix} propose RegMix to automatically identify a high-performing data mixture by formulating it as a regression task.
\citet{li2025data} frame the selection of data proportions as an optimization problem to minimize validation loss across multiple tasks during SFT.

While these studies provide insights into domain-level mixing (e.g., balancing math, code, and chat), they lack an examination of the internal interference between different response formats within the same domain, which is a key focus of our study.
% While these studies provide insights for domain-level mixing (e.g., balancing math, code, and chat), there lack examine the internal interference between different response formats within the same domain, which is a key focus of our study.
% \citet{li2025data} frame data mixing as an optimization problem and introduce a novel method designed to minimize validation loss. 

% However, these studies focus on domain-level proportions.

% Safety tax~\citep{huang2025safetytaxsafetyalignment} shows that there exists a trade-off between reasoning and safety capability with the sequential LRM production pipeline. 

% Qwen3 is an open-weight family of multilingual LLMs spanning both dense and Mixture-of-Experts (MoE) architectures (from 0.6B to 235B parameters), designed to improve capability and inference efficiency across general tasks, coding, and mathematical reasoning~\citep{yang2025qwen3technicalreport}. A key characteristic of Qwen3 is that it natively supports both fast, instruction-following responses and multi-step reasoning within a single model, reducing the need to switch between separate chat-optimized and reasoning-specific systems.

% Thinking Mode Fusion (TMF) is the post-training strategy in Qwen3 that fuses \emph{thinking} and \emph{non-thinking} modes, enabling explicit mode selection through a unified chat template while keeping the output format consistent. TMF further supports a controllable thinking budget, allowing practitioners to trade off latency and performance by limiting the number of reasoning tokens during inference.

\paragraph{Math Datasets.}
We group math datasets into two categories that align with the thinking (with long reasoning) versus non-thinking (with short solutions) modes considered in this paper.

On the simple data side, a complementary line of work focuses on datasets that admit short solutions or direct answer supervision, which better aligns with the non-thinking mode. Classic benchmarks such as GSM8K~\citep{cobbe2021trainingverifierssolvemath}, which consists of grade-school math problems, are widely used for instruction tuning and evaluation, and are often employed in short-solution settings to encourage concise generation.
Calc-X~\citep{kadlcik-etal-2023-calc} is a collection of simple arithmetic-focused math word problems, comprising over $300{,}000$ curated samples.
\citet{patel-etal-2021-nlp} introduced MAWPS, an elementary-level benchmark dataset featuring concise solutions and final answers.
In this work, we use GSM8K as the non-thinking mode base dataset, as it covers a diverse range of math problems and is widely adopted.

On the long reasoning data side, recent work has released large-scale datasets that explicitly provide long reasoning traces, which are crucial for training and analyzing long-form mathematical reasoning. OpenMathReasoning~\citep{moshkov2025aimo2winningsolutionbuilding} is a generated large-scale math reasoning dataset for training large language models (LLMs), containing $3.2$ million long chain-of-thought (CoT) solutions. OpenMathInstruct-1~\citep{NEURIPS2024_3d5aa9a7} and OpenMathInstruct-2~\citep{toshniwal2025openmathinstruct} provide $1.8$ million and $14$ million generated solutions, respectively, covering a wide range of mathematical difficulty levels.
We also note curated math reasoning benchmarks such as OMNI-MATH~\citep{gao2025omnimath} and LIMO~\citep{ye2025limo}, which are commonly used to evaluate advanced reasoning capabilities. In addition, several datasets emphasize challenging questions paired with chain-of-thought annotations, including Skywork-MathQA~\citep{zeng2024skyworkmathdatascalinglaws}, which contains $2.5$ million question-answer pairs, and NuminaMath~\citep{numina_math_datasets}, which consists of $860,000$ competition-level problems with long-form reasoning. 
For ease of evaluation, we choose OpenMathReasoning since it provides problems with exact and verifiable final answers, unlike many other datasets that include proof-based or open-ended reasoning tasks without checkable solutions. In this work, we sample data from the OpenMathReasoning subset generated by DeepSeek-R1~\citep{deepseekai2025deepseekr1incentivizingreasoningcapability} and select examples with a high pass rate (greater than $0.96$) under the 72B model as the thinking-mode training dataset.

\section{Data Processing}\label{Sec:Data Processing}
As mentioned above~(\Cref{Sec:Related Work}), we select GSM8K~\citep{cobbe2021trainingverifierssolvemath} and OpenMathReasoning~\citep{moshkov2025aimo2winningsolutionbuilding} as the base datasets for non-thinking and thinking modes, respectively, which are in accordance with the Qwen3 mode design for simple and complex problems.

\begin{table}[t]
\caption{Benchmark statistics for thinking and non-thinking training data under different data ratios. \textbf{T} denotes thinking mode data and \textbf{NT} denotes non-thinking mode data. We fix the thinking mode subset size ($N_T = 1500$) and vary the non-thinking subset size ($N_{NT}$) accordingly. \textbf{Mean} denotes the mean length of the reasoning traces.}
\label{tab:table_data}
\centering
\footnotesize
\setlength{\tabcolsep}{0.7mm}{
\begin{tabular}{@{}cccccc@{}}
\toprule
\textbf{Mode}& \textbf{Ratio (T : NT)} & \textbf{Count} & \textbf{Mean}& \textbf{Min}& \textbf{Max} \\ 
\midrule
\multirow{7}{*}{\textbf{Non-thinking}} 
& 4 : 1                     & 375            &  671&  330&  1{,}540         \\
& 3 : 1                     & 500            &  672&  330&  1{,}540           \\
& 2 : 1                     & 750            &  681&  330&  1{,}540           \\
& 1 : 1                     & 1{,}500        &  692&  298&  1{,}817          \\
& 1 : 2                     & 3{,}000        &  692&  298&  1{,}817           \\
& 1 : 3                     & 4{,}500        &  695&  298&  1{,}863           \\
& 1 : 4                     & 6{,}000        &  694&  298&  1{,}863          \\ 
\midrule
\textbf{Thinking}                      
& -                       & 1{,}500        &   11{,}877&  1{,}903&  54{,}786          \\ \bottomrule
\end{tabular}}
\end{table}

\paragraph{Non-thinking mode dataset.}
For the non-thinking mode, we use the GSM8K training set. For each experimental setting, we subsample GSM8K into a subset of size $N_{NT}$ according to a target ratio $\rho$ between thinking ($N_T$) and non-thinking ($N_{NT}$) data. We define the ratio as
\begin{equation}
\small
\setlength{\thinmuskip}{1mu}
\setlength{\medmuskip}{1mu}
\rho = \frac{N_T}{N_{NT}} .
\end{equation}
We sweep $\rho$ from $4$ to $1/4$ (i.e., non-thinking to thinking ratios from $4{:}1$ to $1{:}4$). 
% Since the GSM8K training set contains only $7{,}470$ examples, 
% to satisfy these ratio constraints we fix the amount of thinking mode data to $1{,}500$ examples.
Since the GSM8K training set contains only $7{,}470$ examples, to satisfy our experimental design for the data ratio, we fix the amount of thinking mode data to $1{,}500$ examples.
With the thinking mode training data fixed at $1{,}500$, the corresponding non-thinking training data size ranges from $375$ to $6{,}000$. 
\paragraph{Thinking mode dataset.}
To enable controlled comparisons across different fusion strategies and data ratios, we fix the number of thinking mode training examples to $N_T = 1{,}500$ throughout our benchmark. The OpenMathReasoning dataset contains $3.2$ million long chain-of-thought solutions; from this corpus, we sample $1{,}875$ examples. We then randomly split these samples into a $8{:}2$ train/test partition. Detailed statistics of the thinking and non-thinking training data are presented in~\Cref{tab:table_data}.

\begin{table}[t]
\caption{Statistics of test data for thinking and non-thinking evaluation. \textbf{Mean} denotes the mean length of the reasoning traces, \textbf{Min} denotes the minimum length of the reasoning traces, and \textbf{Max} denotes the maximum length of the reasoning traces.}
\label{tab:test_data_stats}
\centering
\footnotesize
\begin{tabular}{@{}ccccc@{}}
\toprule
\textbf{Mode}         & \textbf{Count} & \textbf{Mean} & \textbf{Min} & \textbf{Max} \\ \midrule
\textbf{Non-thinking} & 375            & 696           & 298          & 1{,}863      \\
\textbf{Thinking}     & 375            & 11{,}992      & 1{,}840      & 69{,}268     \\ \bottomrule
\end{tabular}
\end{table}

For uniform evaluation, we sample $375$ test examples from the GSM8K and OpenMathReasoning test sets for the non-thinking and thinking modes, respectively. 
Detailed statistics of the thinking and non-thinking test data are presented in~\Cref{tab:test_data_stats}.
The thinking mode test data have a mean length of $11{,}992$, with a maximum length of $69{,}268$ and a minimum length of $1{,}840$. In contrast, the non-thinking mode test data have a mean length of $696$, with a maximum length of $1{,}863$ and a minimum length of $298$.

After data selection, to align with the Qwen3 official chat template settings, we further process the GSM8K subset into non-thinking mode data and the OpenMathReasoning subset into thinking mode data. For non-thinking mode data, we insert the \texttt{<think></think>} tags before the solutions. For thinking mode data, we wrap the long-form reasoning solutions with the \texttt{<think></think>} tags. To simplify the evaluation process, we wrap all final answers in the \texttt{\textbackslash boxed\{\}} tag. Examples of non-thinking and thinking data are shown in Appendix~\ref{App:Prompting and Mode Formattin}.
 % in~\Cref{App:Example of Data}

\section{Methodology}\label{Sec:Methodology}

\begin{figure*}[th]
\begin{center}
\includegraphics[width=0.75\linewidth]{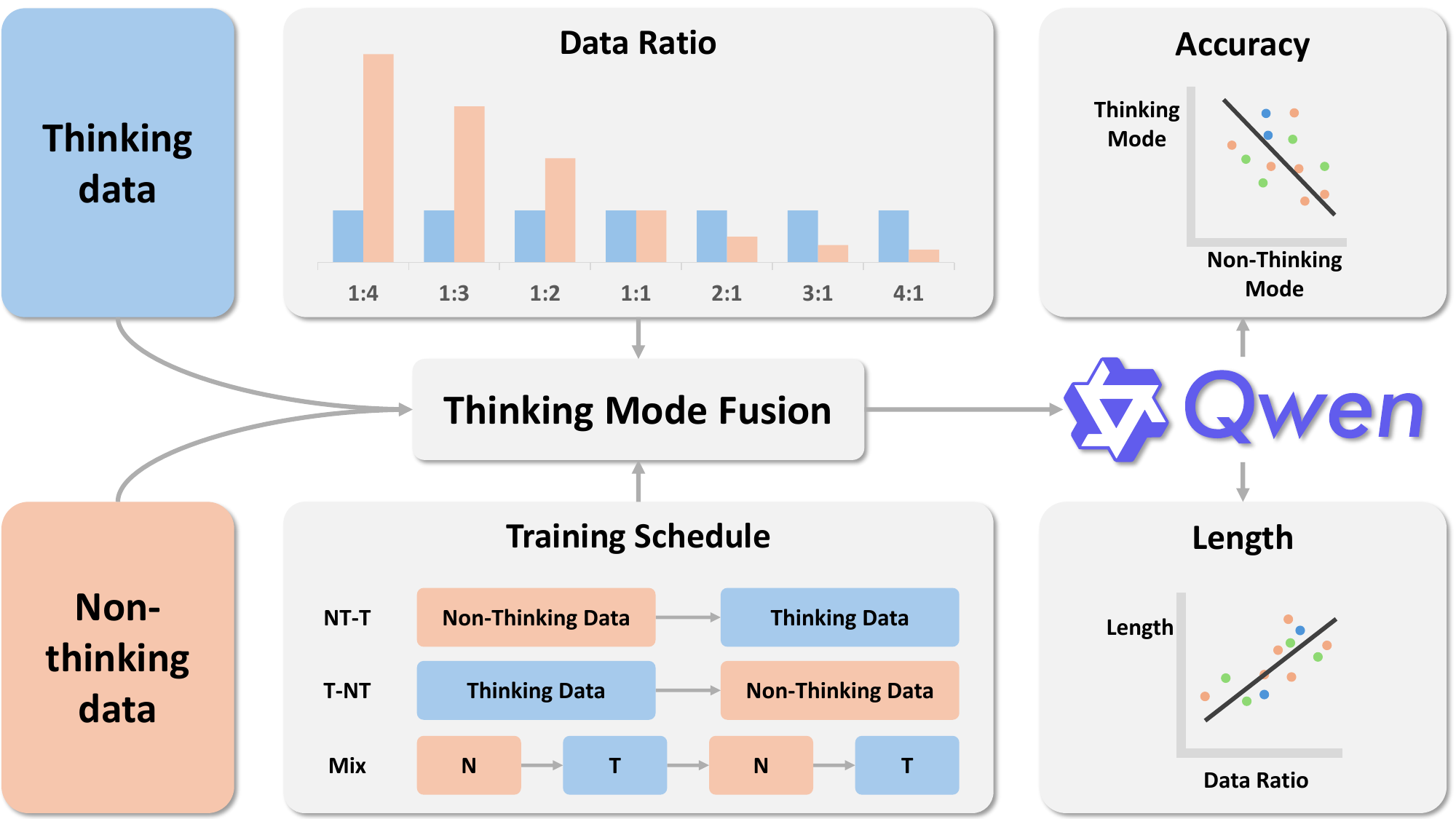}
\caption{Overview of Thinking Mode Fusion training design. We construct two datasets: thinking mode and non-thinking mode. We vary the data ratio (T:NT from $1{:}4$ to $4{:}1$) and schedule the resulting SFT stream using three strategies: \textbf{T-NT} (first thinking then non-thinking), \textbf{NT-T} (non-thinking then thinking), and \textbf{Mix} (interleaved to match the target ratio). Each configuration fine-tunes Qwen3 model and is evaluated on both thinking and non-thinking benchmarks to quantify interference and trade-offs.}
\label{fig:overview}
\end{center}
\end{figure*}

We study Thinking Mode Fusion (TMF) as a post-training problem for large language models (Qwen3 models), where the model support two distinct modes: (i) long-form reasoning for hard problems and (ii) concise responses for simple problems. Our goal is to evaluate how TMF is affected by two controllable factors during supervised fine-tuning (SFT): \textbf{(a) training schedule} (the ordering or interleaving of thinking and non-thinking mode training) and \textbf{(b) data ratio} (the relative amount of thinking versus non-thinking data, ranging from $1{:}4$ to $4{:}1$, while fixing the amount of thinking mode training data). We use the non-thinking and thinking test sets constructed in~\Cref{Sec:Data Processing} to evaluate the uniformity of model performance across modes. \Cref{fig:overview} provides an overview of our experimental design.

\subsection{Supervised Fine-Tuning with Two Modes}
% \subsection{Supervised Fine-Tuning with Two Modes}
We construct two types of training data (thinking and non-thinking) and feed them into a unified supervised fine-tuning (SFT) framework with different training schedules and data ratios. 

As mentioned in~\Cref{Sec:Data Processing}, we convert examples into the model's chat template, with the mode specified by the presence or absence of reasoning content within the \texttt{<think></think>} tags. For thinking-mode training, we use the constructed thinking-mode training data, in which long-form reasoning solutions are wrapped in the \texttt{<think></think>} tags. For non-thinking-mode training, we use the constructed non-thinking-mode training data, which contain short solutions following the \texttt{<think></think>} tags. 

After training, we extract final answers from the \texttt{\textbackslash boxed\{\}} tag, as described in~\Cref{Sec:Data Processing}. Finally, we evaluate model performance in both thinking and non-thinking modes using the corresponding test sets in~\Cref{tab:test_data_stats}.
% TMF enables a single model to support both thinking and non-thinking modes. We construct two types of training data and feed them into a unified supervised fine-tuning (SFT) framework with different training schedules and data ratios. 

% For thinking mode training, we use the constructed thinking mode training data, in which long-form reasoning solutions are wrapped in the \texttt{<think></think>} tags. For non-thinking mode training, we use the constructed non-thinking mode training data, which contain short solutions following the \texttt{<think></think>} tags. 

% To ensure that both modes share a compatible interface, we convert all examples into a unified chat-style format using the model's chat template, with the mode implicitly specified by the presence or absence of reasoning content within the \texttt{<think></think>} tags. After training, we extract final answers from the \texttt{\textbackslash boxed\{\}} tag, as described in~\Cref{Sec:Data Processing}. Finally, we evaluate model performance in both thinking and non-thinking modes using the corresponding test sets in~\Cref{tab:test_data_stats}.

\subsection{Data Ratio Control}
As described in~\Cref{Sec:Data Processing}, we construct thinking mode and non-thinking mode training data with a wide range of ratios $\{1{:}4{,} 1{:}3{,} 1{:}2{,} 1{:}1{,} 2{:}1{,} 3{:}1{,} 4{:}1\}$ by fixing the number of thinking mode training examples.

% Let $\mathcal{D}_T$ denote the thinking mode training set and $\mathcal{D}_{NT}$ denote the non-thinking mode training set. We fix the number of thinking mode training examples to $N_T = 1500$ to control the overall amount of long reasoning supervision. We then vary the size of the non-thinking subset $N_{NT}$ to realize different ratios between the two modes. Following our implementation, we define
% \begin{equation}
% \small
% \setlength{\thinmuskip}{1mu}
% \setlength{\medmuskip}{1mu}
% \rho = \frac{N_T}{N_{NT}},
% \end{equation}
% and sweep $\rho$ from $4$ to $1/4$, corresponding to ratios from $4{:}1$ to $1{:}4$ (T:NT). With $N_T$ fixed, this yields $N_{NT}\in\{375, 500, 750, 1500, 3000, 4500, 6000\}$ (\Cref{tab:table_data}).

% \subsection{Training Schedule for Thinking Mode Fusion}\label{Sec:Training Schedule for Thinking Mode Fusion}
\subsection{Training Schedule}\label{Sec:Training Schedule for Thinking Mode Fusion}
Given a fixed thinking set $\mathcal{D}_T$ and a ratio-specific non-thinking subset $\mathcal{D}_{NT}^{(\rho)}$, TMF training reduces to constructing a single SFT stream $\mathcal{S}$ by a scheduling function
\begin{equation}
\small
\setlength{\thinmuskip}{1mu}
\setlength{\medmuskip}{1mu}
\mathcal{S} = \textsc{Schedule}\big(\mathcal{D}_T, \mathcal{D}_{NT}^{(\rho)}; m\big),
\end{equation}
where $m$ specifies how the two modes are arranged during training. We investigate three schedules that cover sequential curricula and interleaving.

\paragraph{Sequential: \texttt{T-NT}.}
We place all thinking mode examples before all non-thinking mode examples:
\begin{equation}
\small
\setlength{\thinmuskip}{1mu}
\setlength{\medmuskip}{1mu}
\mathcal{S}_{\texttt{T-NT}} = [\mathcal{D}_T \;\Vert\; \mathcal{D}_{NT}^{(\rho)}].
\end{equation}
This schedule tests whether updates from non-thinking data applied after reasoning training dilute or overwrite reasoning behaviors, and whether reasoning training provides transferable benefits to subsequent concise answering.

\paragraph{Sequential: \texttt{NT-T}.}
We reverse the order:
\begin{equation}
\small
\setlength{\thinmuskip}{1mu}
\setlength{\medmuskip}{1mu}
\mathcal{S}_{\texttt{NT-T}} = [\mathcal{D}_{NT}^{(\rho)} \;\Vert\; \mathcal{D}_T].
\end{equation}
This schedule investigates the opposite direction of interference: whether late-stage reasoning supervision disrupts concise response patterns learned from non-thinking data and whether non-thinking alignment persists under subsequent long-reasoning updates.

\paragraph{Interleaved: \texttt{Mix}.}
We interleave the two modes of data to the target ratio during training. In particular, we use a simple 1-to-$n$ interleaving procedure: when $\rho \le 1$ (thinking is less frequent), we interleave \textbf{1 T (thinking mode)} followed by \textbf{$n$ NT (non-thinking mode)} examples, with $n \approx \mathrm{round}(1/\rho)$; when $\rho > 1$ (thinking is more frequent), we interleave \textbf{1 NT} followed by \textbf{$n$ T} examples, with $n \approx \mathrm{round}(\rho)$. This schedule makes the intended TMF usage where the model repeatedly alternates between the two modes, enabling us to test whether frequent mode switching alleviates the degradation observed under purely sequential curricula.

\subsection{Full Factorial Design}
Our TMF analysis is defined by the Cartesian product of data ratios and schedules:
% \begin{equation}
% (\rho, m)\in \{4, 3, 2, 1, \tfrac{1}{2}, \tfrac{1}{3}, \tfrac{1}{4}\}\times\{\texttt{T-NT},\texttt{NT-T},\texttt{Mix}\}.
% \end{equation}
\begin{equation}
\small
\setlength{\thinmuskip}{1mu}
\setlength{\medmuskip}{1mu}
(\rho, m)\in \{4, 3, 2, 1, \tfrac{1}{2}, \tfrac{1}{3}, \tfrac{1}{4}\}
\times\{\texttt{T-NT},\texttt{NT-T},\texttt{Mix}\}.
\end{equation}
For each configuration $(\rho,m)$, we build the corresponding SFT training $\mathcal{S}$ while keeping the base model, supervision format, and optimization hyperparameters fixed. This controlled setup allows us to attribute performance changes to (i) how much thinking supervision is mixed with non-thinking data and (ii) how the two modes are scheduled during training, thereby directly quantifying interference and trade-offs induced by TMF. We finally evaluate both the accuracy and the reasoning length of model responses, as reasoning length has been shown to reflect reasoning quality in mathematical problem solving~\citep{sprague2025to,jin-etal-2024-impact,tutek-etal-2025-measuring}.

\section{Experiment and Analysis}

\subsection{Experiment Setup}\label{Sec:Experiment}

Our study aims to answer four research questions:

\textbf{RQ1: Do the non-thinking and thinking modes affect each other's performance?}

Although Qwen3 adopts Thinking Mode Fusion to separate long-form and concise reasoning into different fine-tuning modes, training these two modes jointly may still lead to performance trade-offs or interference in the mathematical domain.

\textbf{RQ2: How does the different training schedule affect Thinking Mode Fusion?}  
Thinking Mode Fusion is introduced by first fine-tuning on concise reasoning traces and then fine-tuning on long-form reasoning traces~\citep{yang2025qwen3technicalreport}. However, in practice, non-thinking and thinking modes can be trained in different orders. As described in~\Cref{Sec:Training Schedule for Thinking Mode Fusion}, we consider three training schedules, \{T-NT, NT-T, Mix\}, and evaluate model performance under different training schedules.

\textbf{RQ3: How does the different data ratio affect Thinking Mode Fusion?}  
In real-world applications, short-form reasoning data are far more abundant than high-quality long-form reasoning data. Therefore, the data ratio between thinking and non-thinking modes is crucial for understanding the effectiveness of Thinking Mode Fusion.

\textbf{RQ4: What is the quantitative relationship between the non-thinking and thinking modes?} 
After analyzing the effects of training schedules and data ratios, it is essential to quantify the relationship between the two modes to better guide TMF training for LLMs.

All detailed experimental settings and baselines can be found in Appendix~\ref{App:training setting}.

\begin{table*}[t]
\caption{Performance comparison of accuracy and reasoning length under different data ratios and training schedule settings.
% T-NT denotes a schedule that places all thinking mode examples before all non-thinking mode examples while NT-T denotes reverse the order. Mix denotes a schedule that interleave the two modes of data to approximate the target ratio during training.
We report Len (average response length) and Acc (accuracy) for three runs. The rightmost Mean column averages results across all data ratios, while the Mean row in each mode block averages over training schedules. The Mix training schedule achieves the highest overall accuracy, while increasing the thinking data ratio consistently improves thinking-mode performance and leads to longer responses.}
\label{tab:all_comparison}
\centering
\scriptsize
\setlength{\tabcolsep}{0.7mm}{
\begin{tabular}{@{}cccccccccccccccccc@{}}
\toprule
                                        &                  & \multicolumn{14}{c}{\textbf{Data Ratio (T:NT, T=1500)}}                                                                                                                                                                                                                                                                                                                                                                                                                             & \multicolumn{2}{c}{}                                           \\ \cmidrule(lr){3-16}
\textbf{Mode}                           & \textbf{Setting} & \multicolumn{2}{c}{\textbf{1:4}}                               & \multicolumn{2}{c}{\textbf{1:3}}                               & \multicolumn{2}{c}{\textbf{1:2}}                               & \multicolumn{2}{c}{\textbf{1:1}}                               & \multicolumn{2}{c}{\textbf{2:1}}                               & \multicolumn{2}{c}{\textbf{3:1}}                               & \multicolumn{2}{c}{\textbf{4:1}}                               & \multicolumn{2}{c}{\multirow{-2}{*}{\textbf{Mean}}}            \\ \cmidrule(l){3-18} 
                                        &                  & \textbf{Len}                   & \textbf{Acc}                  & \textbf{Len}                   & \textbf{Acc}                  & \textbf{Len}                   & \textbf{Acc}                  & \textbf{Len}                   & \textbf{Acc}                  & \textbf{Len}                   & \textbf{Acc}                  & \textbf{Len}                   & \textbf{Acc}                  & \textbf{Len}                   & \textbf{Acc}                  & \textbf{Len}                   & \textbf{Acc}                  \\ \midrule
                                        & T-NT             & 271.3                          & 0.737                         & 245.9                          & 0.692                         & 254.8                          & 0.700                         & 246.8                          & 0.640                         & 316.6                          & 0.695                         & 295.1                          & 0.667                         & 328.8                          & 0.676                         & \cellcolor[HTML]{DAE8FC}279.9  & \cellcolor[HTML]{DAE8FC}0.687 \\
                                        & NT-T             & 245.1                          & 0.664                         & 260.1                          & 0.716                         & 278.2                          & 0.684                         & 270.3                          & 0.692                         & 269.1                          & 0.696                         & 343.6                          & 0.697                         & 399.8                          & 0.693                         & \cellcolor[HTML]{DAE8FC}295.2  & \cellcolor[HTML]{DAE8FC}0.692 \\
                                        & Mix              & 268.3                          & 0.703                         & 254.3                          & 0.706                         & 255.1                          & 0.676                         & 272.1                          & 0.717                         & 273.7                          & 0.698                         & 326.7                          & 0.677                         & 360.5                          & 0.687                         & \cellcolor[HTML]{DAE8FC}287.2  & \cellcolor[HTML]{DAE8FC}\underline{\textbf{0.695}} \\
\multirow{-4}{*}{\textbf{Non-thinking}} & Mean             & \cellcolor[HTML]{DAE8FC}261.6  & \cellcolor[HTML]{DAE8FC}0.701 & \cellcolor[HTML]{DAE8FC}253.4  & \cellcolor[HTML]{DAE8FC}\underline{\textbf{0.705}} & \cellcolor[HTML]{DAE8FC}262.7  & \cellcolor[HTML]{DAE8FC}0.687 & \cellcolor[HTML]{DAE8FC}263.1  & \cellcolor[HTML]{DAE8FC}0.683 & \cellcolor[HTML]{DAE8FC}286.5  & \cellcolor[HTML]{DAE8FC}0.696 & \cellcolor[HTML]{DAE8FC}321.8  & \cellcolor[HTML]{DAE8FC}0.680 & \cellcolor[HTML]{DAE8FC}363.0  & \cellcolor[HTML]{DAE8FC}0.685 & \cellcolor[HTML]{DAE8FC}--     & \cellcolor[HTML]{DAE8FC}--    \\ \midrule
                                        & T-NT             & 5455.0                         & 0.188                         & 5632.5                         & 0.198                         & 5065.0                         & 0.174                         & 5519.8                         & 0.186                         & 6242.4                         & 0.220                         & 6858.5                         & 0.224                         & 7027.6                         & 0.221                         & \cellcolor[HTML]{DAE8FC}5971.5 & \cellcolor[HTML]{DAE8FC}0.202 \\
                                        & NT-T             & 5731.7                         & 0.200                         & 5779.5                         & 0.192                         & 5938.7                         & 0.204                         & 5606.7                         & 0.196                         & 6225.7                         & 0.212                         & 6682.2                         & 0.224                         & 7080.8                         & 0.227                         & \cellcolor[HTML]{DAE8FC}6149.3 & \cellcolor[HTML]{DAE8FC}\underline{\textbf{0.208}} \\
                                        & Mix              & 5877.1                         & 0.208                         & 5333.3                         & 0.204                         & 5438.8                         & 0.190                         & 5911.4                         & 0.193                         & 6211.7                         & 0.215                         & 7000.4                         & 0.218                         & 6888.5                         & 0.225                         & \cellcolor[HTML]{DAE8FC}6094.5 & \cellcolor[HTML]{DAE8FC}\underline{\textbf{0.208}} \\
\multirow{-4}{*}{\textbf{Thinking}}     & Mean             & \cellcolor[HTML]{DAE8FC}5687.9 & \cellcolor[HTML]{DAE8FC}0.199 & \cellcolor[HTML]{DAE8FC}5581.8 & \cellcolor[HTML]{DAE8FC}0.198 & \cellcolor[HTML]{DAE8FC}5480.8 & \cellcolor[HTML]{DAE8FC}0.189 & \cellcolor[HTML]{DAE8FC}5679.3 & \cellcolor[HTML]{DAE8FC}0.192 & \cellcolor[HTML]{DAE8FC}6226.6 & \cellcolor[HTML]{DAE8FC}0.216 & \cellcolor[HTML]{DAE8FC}6847.0 & \cellcolor[HTML]{DAE8FC}0.222 & \cellcolor[HTML]{DAE8FC}6999.0 & \cellcolor[HTML]{DAE8FC}\underline{\textbf{0.224}} & \cellcolor[HTML]{DAE8FC}--     & \cellcolor[HTML]{DAE8FC}--    \\ \bottomrule
\end{tabular}}
\end{table*}

% \subsection{Do the non-thinking and thinking modes affect each other’s performance?}
\subsection{RQ1: Interaction between Non-Thinking and Thinking Mode}
% Our results reveal an asymmetric interaction between the two modes: increasing the ratio of non-thinking supervision reduces the accuracy of the thinking mode.
The accuracy and reasoning length performance on the test set under training with different data ratios and training schedule settings are presented in~\Cref{tab:all_comparison}. 
Rows correspond to three schedules with two modes: T-NT (thinking before non-thinking), NT-T (non-thinking before thinking), and Mix (interleaved training). Columns are organized by the thinking to non-thinking (T:NT) data ratio from $1{:}4$ to $4{:}1$, reporting Len (average response length) and Acc (accuracy) for three runs. The rightmost Mean column averages results across all data ratios, while the Mean row in each mode block averages over training schedules. 

The results indicate that the non-thinking and thinking modes interact during Thinking Mode Fusion (TMF) training. When averaging results across all data ratios, the Mix training schedule achieves the highest accuracy on both the thinking and non-thinking test sets, with accuracies of $0.208$ and $0.695$, respectively.
In thinking-mode evaluation, the NT-T training schedule also attains the highest accuracy while producing the longest average response length. When averaging results across all training schedules, the thinking mode achieves its highest accuracy at the largest thinking data ratio (T:NT=$4{:}1$), whereas the non-thinking mode achieves its highest accuracy at the second highest non-thinking data ratio (T:NT=$1{:}3$).

Both thinking and non-thinking modes produce the longest response length at the largest thinking data ratio (T:NT=$4{:}1$) when averaging over the training schedules. Similarly, both modes produce the longest response under the NT-T training schedule when averaging over the data ratio. These results are consistent with the intuition and prior work that tuning with long-form data at the end tends to yield longer responses~\citep{10.5555/3692070.3694581,koksal-etal-2024-longform}.

Overall, these results demonstrate that the performance of TMF is affected by the interaction between thinking and non-thinking modes, as modulated by variations in the data ratio and the training schedule. The Mix training schedule achieves the highest overall accuracy, while increasing the thinking data ratio consistently improves thinking-mode performance and leads to longer responses.
\subsection{RQ2: Effect of Training Schedule}

\begin{figure}[t]
\begin{center}
\subfigure[Accuracy on non-thinking and thinking mode test sets after TMF.]{\includegraphics[width=0.75\linewidth]{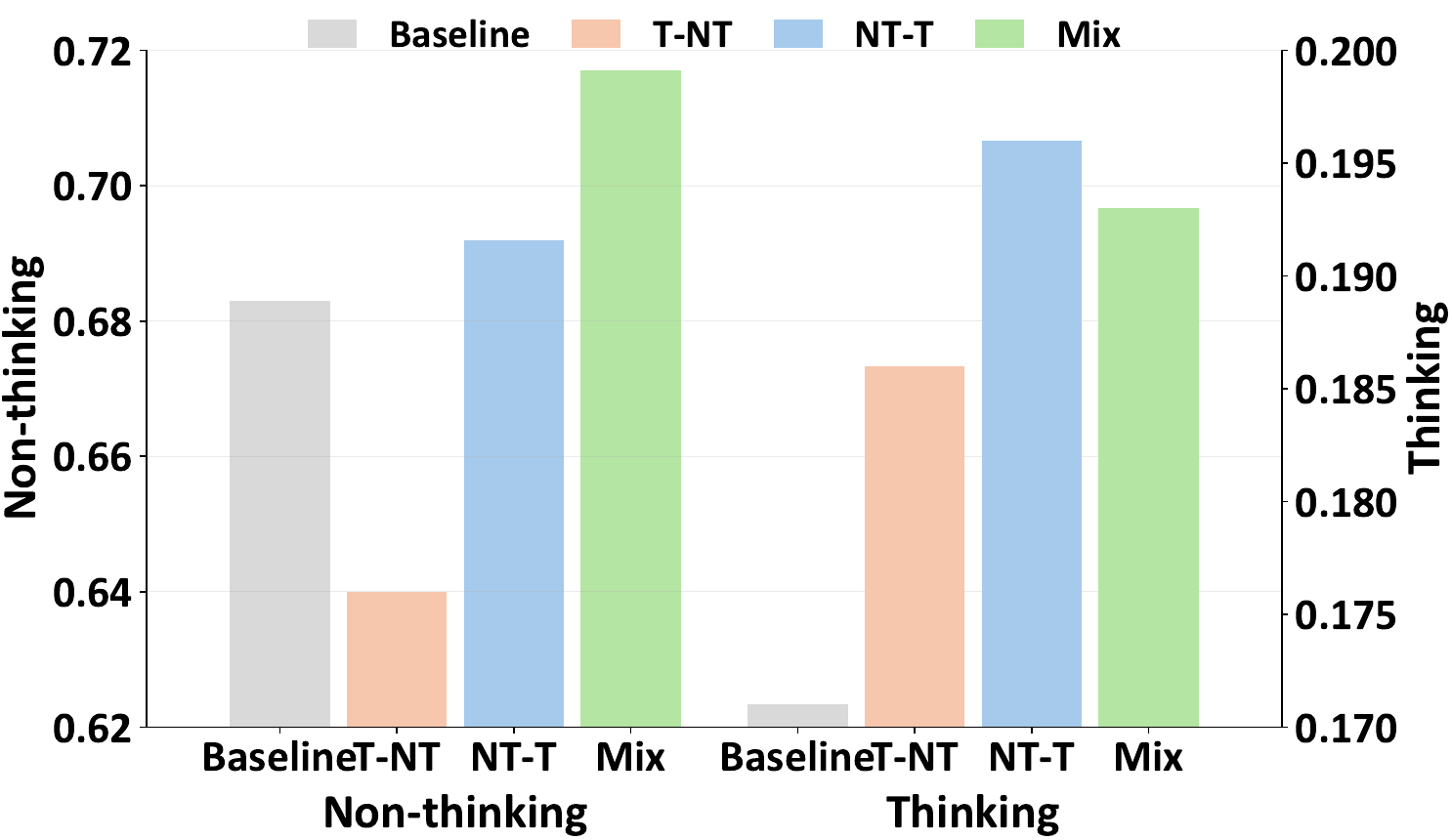}}
\subfigure[Response length on non-thinking and thinking mode test sets after TMF.]{\includegraphics[width=0.75\linewidth]{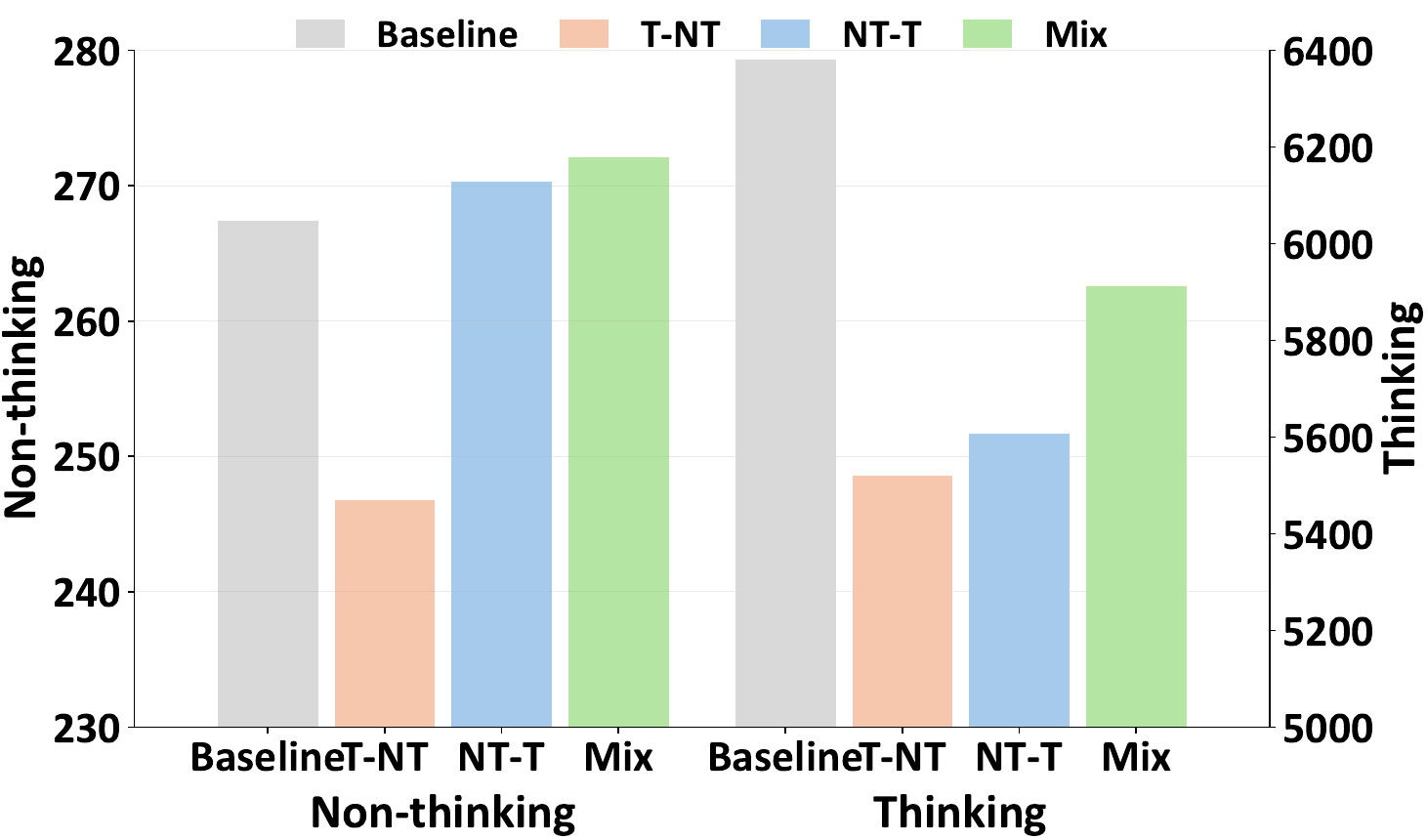}}
\caption{The effect of different training schedules under a fixed thinking to non-thinking data ratio of $1{:}1$.}
\label{fig:training schedule}
\end{center}
\end{figure}

% As presented in~\Cref{fig:training schedule}, we zoom in on a data ratio of 1:1 to explore the effect of different training schedules. 
% We use the performance of our test set of the base model without tuning as the baseline, showing the gray bar. We use the orange, blue, green bar show the performance under T-NT, NT-T and Mix training schedules, respectively.
% Different training schedules have a substantial influence on Thinking Mode Fusion training.
As shown in~\Cref{fig:training schedule}, when fixing the data ratio at 1:1, the results indicate that the choice of training schedule has a substantial impact on Thinking Mode Fusion training.
The performance of the untuned base model on the test set is used as a baseline and is indicated by the gray bar.
The orange, blue, and green bars represent the performance under the T-NT, NT-T, and Mix training schedules, respectively.

In terms of the non-thinking mode, the Mix schedule achieves the highest accuracy, while the T-NT schedule yields the lowest accuracy. Moreover, the T-NT schedule shows a degradation in performance, as the base model already achieves an accuracy of $0.683$ without tuning. The response length exhibits the same pattern as accuracy.

In terms of thinking mode, the NT-T schedule achieves the highest accuracy, while the T-NT schedule results in the lowest accuracy. Compared with the performance of the base model, all training schedules lead to accuracy improvements. The Mix schedule produces the longest response length, which differs from the trend observed in the non-thinking mode.

% \subsection{How does the data ratio affect Thinking Mode Fusion?}
\subsection{RQ3: Effect of Data Ratio}

% \begin{figure}[t]

\begin{figure*}[t]
\begin{center}
\subfigure[Non-thinking accuracy]{\includegraphics[width=0.4\linewidth]{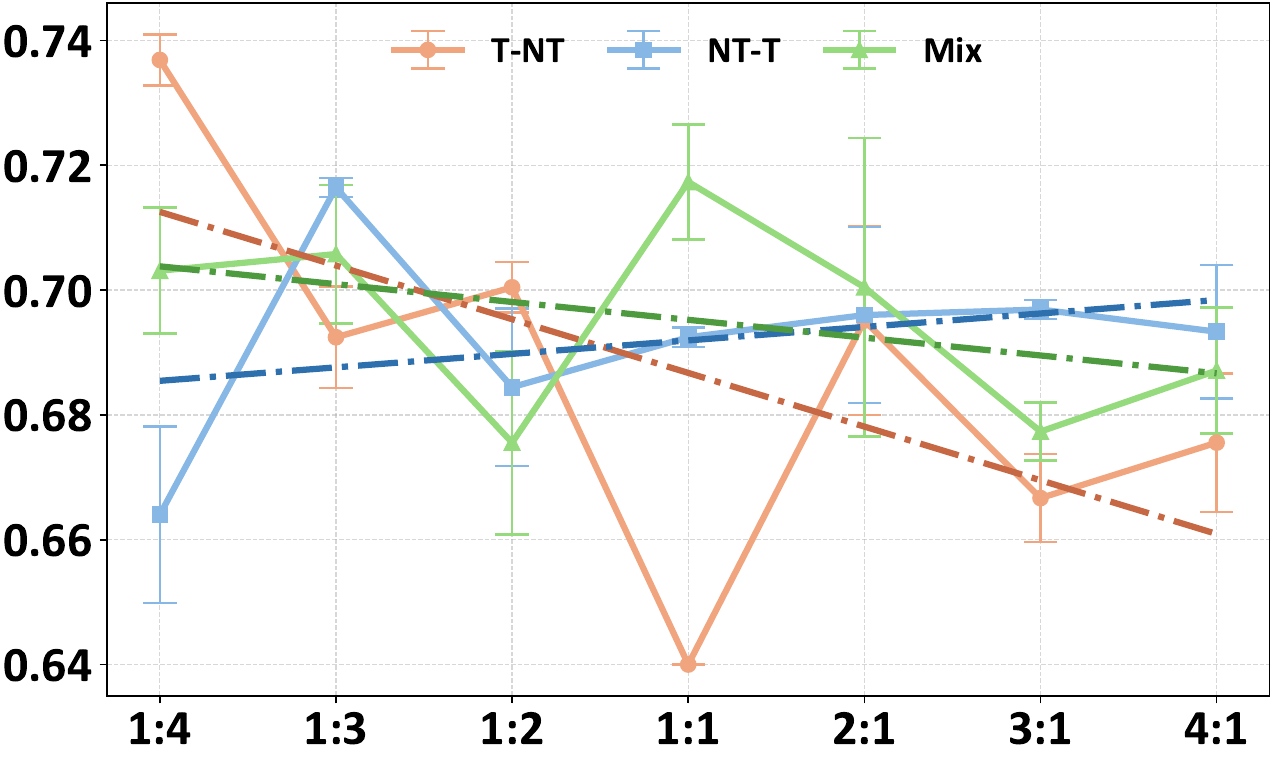}}
\subfigure[Non-thinking length]{\includegraphics[width=0.4\linewidth]{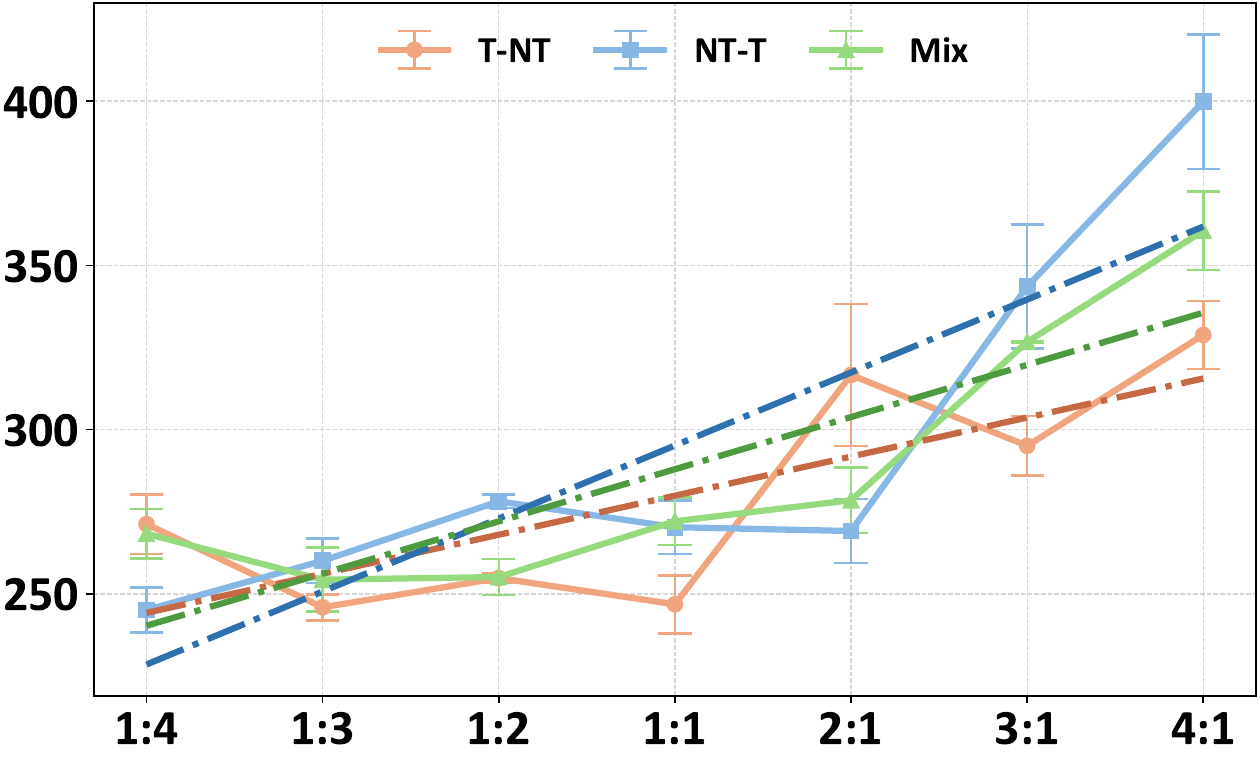}}
\subfigure[Thinking accuracy]{\includegraphics[width=0.4\linewidth]{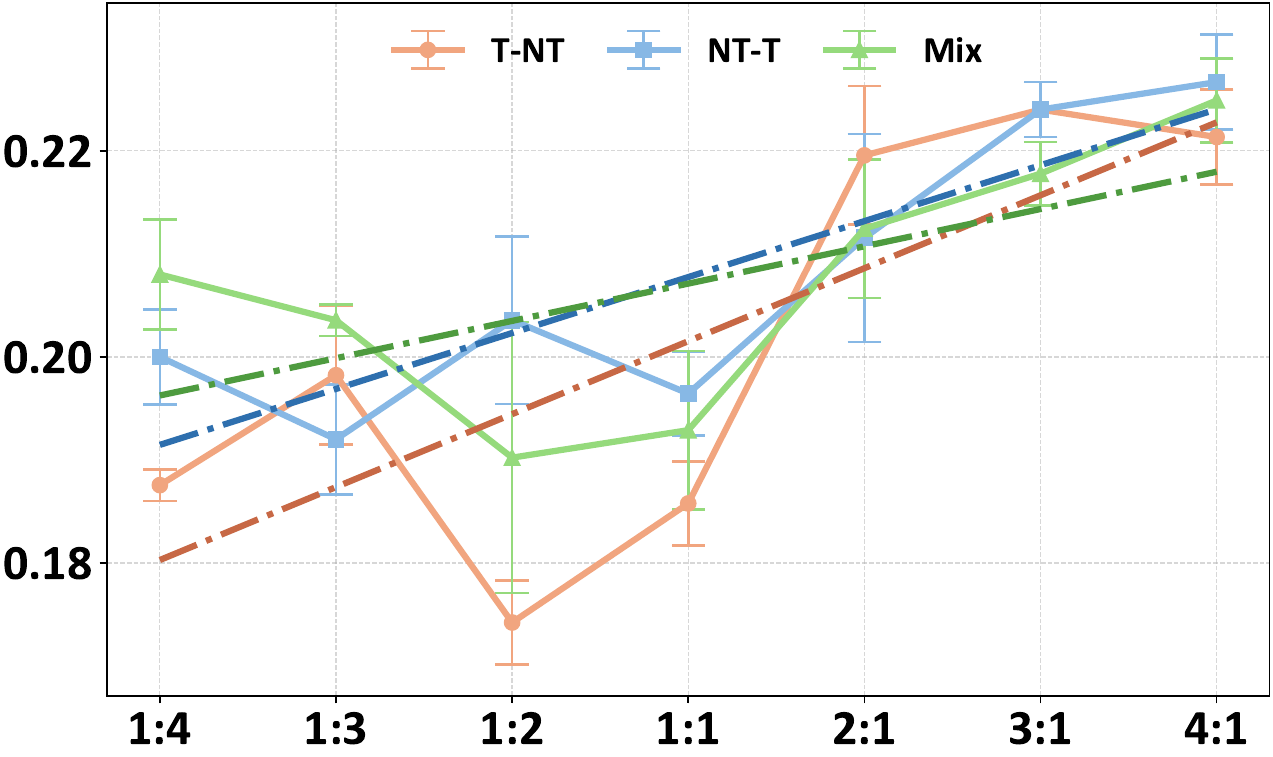}}
\subfigure[Thinking length]{\includegraphics[width=0.4\linewidth]{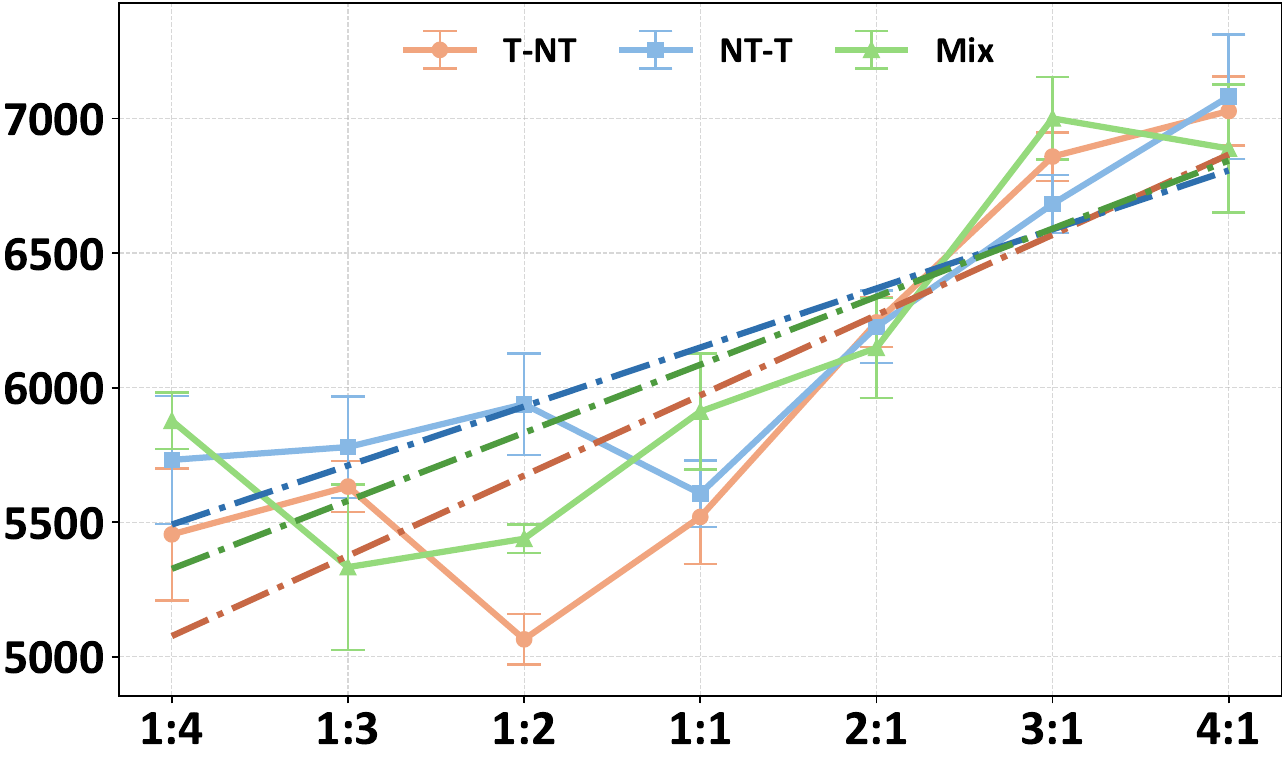}}
\caption{Performance comparison of accuracy and reasoning length under different data ratios and training schedule settings on non-thinking mode, averaged over three runs. Higher non-thinking ratios improve non-thinking accuracy but degrade thinking accuracy; under the Mix schedule, non-thinking accuracy increases with the data ratio, while thinking accuracy consistently increases across all schedules.}
\label{fig:fig4}
\end{center}
\end{figure*}

The trend comparison of accuracy and reasoning length under different data ratios and training schedules is presented in~\Cref{fig:fig4}. 
The orange, blue, and green lines represent the performance under the T-NT, NT-T, and Mix training schedules, respectively. The dashed lines indicate fitted trend lines.

The accuracy of the non-thinking mode shows a decreasing trend as the data ratio increases under the T-NT and NT-T training schedules, whereas it shows an increasing trend as the data ratio increases under the Mix training schedule. However, performance is sensitive to the choice of training schedule under the same data ratio. The T-NT schedule achieves the highest performance when the data ratio is $1{:}4$, while the Mix schedule achieves the highest performance when the data ratio is $4{:}1$. When non-thinking mode data demonstrate an advantage, the T-NT schedule is preferable; when thinking mode data demonstrate an advantage, the Mix schedule is preferable. When the data ratio is $1{:}1$, the Mix schedule yields the best performance. Moreover, the response length of the non-thinking mode shows a consistent increase as the data ratio increases across all training schedules. These results indicate that different data ratios should be paired with different training schedules.

The accuracy of the thinking mode exhibits a consistent increase as the data ratio increases across all training schedules. The Mix schedule achieves the highest accuracy when the data ratio is $1{:}4$, while the NT-T schedule achieves the highest accuracy when the data ratio is $4{:}1$. Notably, the thinking mode training data are fixed at $1{,}500$ examples, indicating that increasing the amount of non-thinking mode data leads to performance degradation in the thinking mode. The reasoning length of the thinking mode also shows a increase as the data ratio increases across all training schedules.

Overall, the data ratio plays a critical role in TMF by systematically trading off non-thinking and thinking performance: higher non-thinking proportions improve concise-answer accuracy but degrade long-form reasoning accuracy. Moreover, the optimal training schedule is strongly coupled with the data ratio.

\subsection{RQ4: The Quantitative Relationship between Two Modes}

 % under different training schedules and data ratios during Thinking Mode Fusion training
\begin{figure}[t]
\centering
\includegraphics[width=0.9\linewidth]{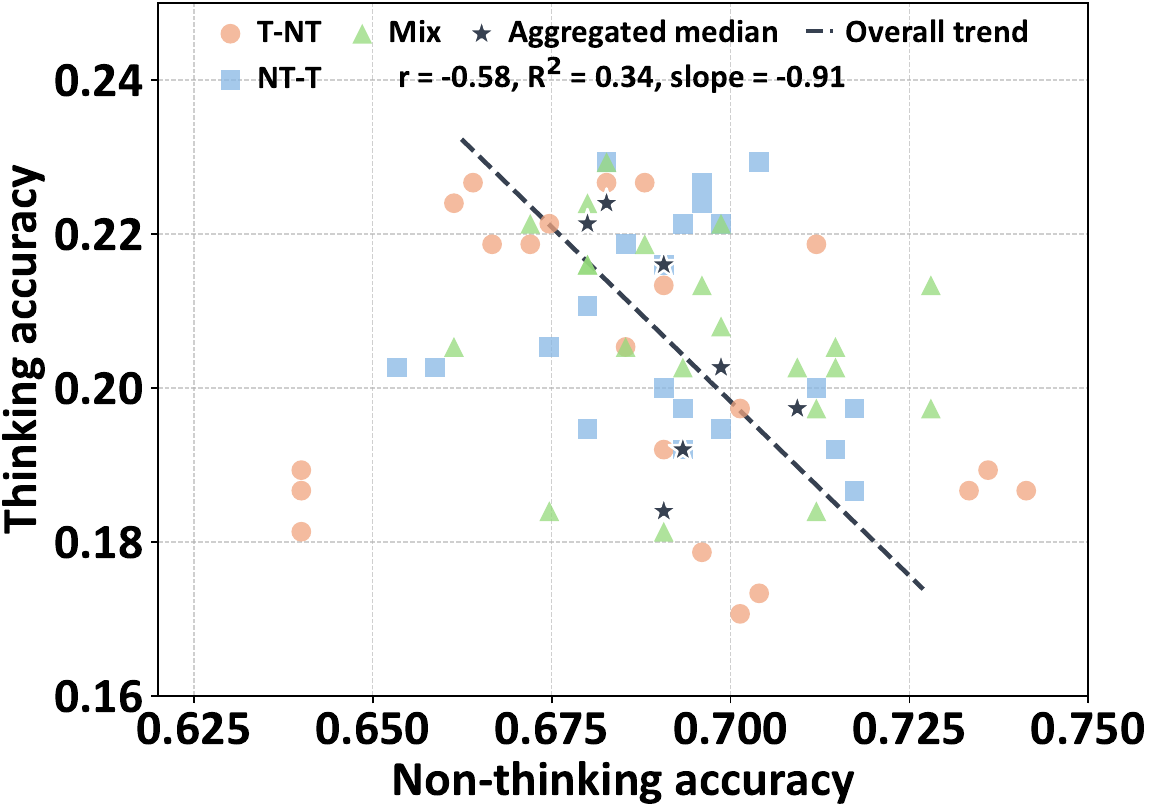}  % 新的图
\caption{Scatter plot of different training schedules and data ratios.
The dashed line indicates the fitted trend line of aggregation of the training schedules.
The line shows a clear negative correlation ($r=-0.58$, $R^2=0.34$), with a regression slope of $-0.91$. 
% The non-thinking and thinking modes exhibit a negative correlation across different data ratios.
}
\label{fig:fig5}
\end{figure}

% revise
We further explore the quantitative relationship between the non-thinking and thinking modes under different training schedules and data ratios, and present a scatter plot in~\Cref{fig:fig5}. The orange, blue, and green scatter points represent accuracy under the T-NT, NT-T, and Mix training schedules, respectively, while the dashed line indicates the fitted trend line of aggregation of the training schedules.

% The dashed line showing a decreasing trend illustrates a clear trade-off: the non-thinking and thinking modes exhibit a negative correlation across different data ratios. 
% The dashed line reveals a negative correlation ($r = -0.58$) between the two modes across different ratio settings.
% % , indicating that higher non-thinking training tends to be associated with lower thinking training across different ratio settings. 
% The coefficient of determination ($R^2 = 0.34$) suggests that approximately 34\% of the variance in thinking accuracy can be explained by its linear relationship with non-thinking accuracy. 
% Furthermore, the estimated regression slope ($\text{slope} = -0.91$) indicates that, on average, an increase of 0.01 in non-thinking accuracy corresponds to a decrease of approximately 0.0091 in thinking accuracy, reflecting a clear trade-off between the two performance measures.
The decreasing trend reveals a clear trade-off between the two modes. Specifically, non-thinking and thinking accuracies exhibit a negative correlation ($r=-0.58$), with a coefficient of determination of $R^2=0.34$. The estimated regression slope ($-0.91$) indicates that a $0.01$ increase in non-thinking mode performance is associated with an average decrease of approximately $0.0091$ in thinking mode performance.

The two modes hold different and potentially competing objectives: the non-thinking mode favors concise responses, whereas the thinking mode emphasizes long-form reasoning. Consequently, encapsulating these conflicting modes induces a negative correlation.

% The chart illustrates a trade-off: The non-thinking and thinking modes exhibit a negative correlation across different training schedules and data ratios, as all dashed lines show a decreasing trend. 
% Particularly, the scatter points of the T-NT training schedule are more widely distributed, and the dashed line is closest to the origin.

% The quantitative relationship between the non-thinking and thinking modes under different training schedules and data ratios during Thinking Mode Fusion training is presented in~\Cref{fig:fig6}. 

% We project the performance of the non-thinking and thinking modes across different training schedules and data ratios onto a Pareto chart to illustrate their trade-offs.
% The non-thinking and thinking modes exhibit a negative correlation across different training schedules and data ratios.

\section{Conclusion}\label{Sec:Conclusion}
In this work, we present a systematic study of the effects of the training schedule and the data ratio in TMF, revealing a trade-off between the two modes in which increased non-thinking supervision degrades thinking mode performance.
Moreover, we show that the training schedule modulates this trade-off: when averaging results across all data ratios, the Mix training schedule achieves the highest accuracy on both the thinking and non-thinking test sets.
Our results show that the optimal training schedule depends on the data ratio, and that the Mix training schedule performs well across varying data ratios for both modes.
Finally, we quantify a negative correlation between non-thinking and thinking mode supervision, highlighting an inherent tension between these two modes. 

% Our results reveal an asymmetric interaction between the two modes: 
% % the trade-off between the two modes shows that thinking mode training results in performance degradation as non-thinking supervision increases.
% the trade-off between the two modes shows that thinking mode performance degrades as non-thinking supervision increases.
% Moreover, we show that the training schedule modulates this trade-off and that the optimal schedule depends on the data ratio.
% Finally, we quantify a negative correlation between non-thinking and thinking mode supervision, highlighting an inherent tension between these two modes. 

% In this work, we presented a systematic study of Thinking Mode Fusion (TMF) as a post-training strategy for unifying concise and long-form reasoning within a single large language model. 

% By jointly varying the training schedule and the data ratio between thinking and non-thinking modes, we analyzed how different TMF configurations affect mathematical problem solving across varying combinations. Our results reveal an asymmetric interaction between the two modes: thinking supervision has a limited impact on non-thinking mode performance, whereas increasing non-thinking supervision slightly degrades thinking accuracy. Moreover, we show that the training order significantly modulates this trade-off, and that the optimal schedule depends on the data ratio. These findings highlight that effective TMF requires careful co-design of data composition and training order, rather than treating multi-mode supervision as a simple mixture.

\section{Limitations}\label{Sec:Limitations}
Our analysis is restricted to mathematical reasoning, and it remains unclear whether the observed interference patterns generalize to other domains, such as code generation. 
Second, we fix the amount of thinking mode data to enable controlled comparisons; exploring regimes in which both thinking and non-thinking data scale jointly may uncover richer dynamics. 
Finally, while we quantify the trade-off between concise answering and long-form reasoning, developing training strategies that explicitly mitigate this tension remains an open challenge.

\section{Ethics Statement}\label{Sec:Ethic}
We do not foresee any particular ethical concerns with our study, which analyzes existing Qwen3 models and is unlikely to lead to unforeseen uses of those models. 
We use two mathematics datasets, GSM8K and OpenMathReasoning, both of which are publicly available. 
The toolkits used in our experiments, including Transformers and TRL, are also publicly available under Apache-2.0 License.

% \section*{Acknowledgments}

\bibliography{custom}

@inproceedings{NEURIPS2022_9d560961,
 author = {Wei, Jason and Wang, Xuezhi and Schuurmans, Dale and Bosma, Maarten and ichter, brian and Xia, Fei and Chi, Ed and Le, Quoc V and Zhou, Denny},
 booktitle = {Advances in Neural Information Processing Systems},
 editor = {S. Koyejo and S. Mohamed and A. Agarwal and D. Belgrave and K. Cho and A. Oh},
 pages = {24824--24837},
 publisher = {Curran Associates, Inc.},
 title = {Chain-of-Thought Prompting Elicits Reasoning in Large Language Models},
 url = {https://proceedings.neurips.cc/paper_files/paper/2022/file/9d5609613524ecf4f15af0f7b31abca4-Paper-Conference.pdf},
 volume = {35},
 year = {2022}
}

@inproceedings{NEURIPS2022_8bb0d291,
 author = {Kojima, Takeshi and Gu, Shixiang (Shane) and Reid, Machel and Matsuo, Yutaka and Iwasawa, Yusuke},
 booktitle = {Advances in Neural Information Processing Systems},
 editor = {S. Koyejo and S. Mohamed and A. Agarwal and D. Belgrave and K. Cho and A. Oh},
 pages = {22199--22213},
 publisher = {Curran Associates, Inc.},
 title = {Large Language Models are Zero-Shot Reasoners},
 url = {https://proceedings.neurips.cc/paper_files/paper/2022/file/8bb0d291acd4acf06ef112099c16f326-Paper-Conference.pdf},
 volume = {35},
 year = {2022}
}

@inproceedings{
wang2023selfconsistency,
title={Self-Consistency Improves Chain of Thought Reasoning in Language Models},
author={Xuezhi Wang and Jason Wei and Dale Schuurmans and Quoc V Le and Ed H. Chi and Sharan Narang and Aakanksha Chowdhery and Denny Zhou},
booktitle={The Eleventh International Conference on Learning Representations },
year={2023},
url={https://openreview.net/forum?id=1PL1NIMMrw}
}

@misc{openai2024openaio1card,
      title={OpenAI o1 System Card}, 
      author={OpenAI},
      year={2024},
      eprint={2412.16720},
      archivePrefix={arXiv},
      primaryClass={cs.AI},
      url={https://arxiv.org/abs/2412.16720}, 
}

@misc{openai2025openaio1card,
      title={GPT-5.1: A smarter, more conversational ChatGPT}, 
      author={OpenAI},
      year={2025},
      primaryClass={cs.AI}, 
      url={https://openai.com/index/gpt-5-1/}
}

@misc{yang2025qwen3technicalreport,
      title={Qwen3 Technical Report}, 
      author={An Yang and Anfeng Li and Baosong Yang and Beichen Zhang and Binyuan Hui and Bo Zheng and Bowen Yu and Chang Gao and Chengen Huang and Chenxu Lv and Chujie Zheng and Dayiheng Liu and Fan Zhou and Fei Huang and Feng Hu and Hao Ge and Haoran Wei and Huan Lin and Jialong Tang and Jian Yang and Jianhong Tu and Jianwei Zhang and Jianxin Yang and Jiaxi Yang and Jing Zhou and Jingren Zhou and Junyang Lin and Kai Dang and Keqin Bao and Kexin Yang and Le Yu and Lianghao Deng and Mei Li and Mingfeng Xue and Mingze Li and Pei Zhang and Peng Wang and Qin Zhu and Rui Men and Ruize Gao and Shixuan Liu and Shuang Luo and Tianhao Li and Tianyi Tang and Wenbiao Yin and Xingzhang Ren and Xinyu Wang and Xinyu Zhang and Xuancheng Ren and Yang Fan and Yang Su and Yichang Zhang and Yinger Zhang and Yu Wan and Yuqiong Liu and Zekun Wang and Zeyu Cui and Zhenru Zhang and Zhipeng Zhou and Zihan Qiu},
      year={2025},
      eprint={2505.09388},
      archivePrefix={arXiv},
      primaryClass={cs.CL},
      url={https://arxiv.org/abs/2505.09388}, 
}

@misc{cobbe2021trainingverifierssolvemath,
      title={Training Verifiers to Solve Math Word Problems}, 
      author={Karl Cobbe and Vineet Kosaraju and Mohammad Bavarian and Mark Chen and Heewoo Jun and Lukasz Kaiser and Matthias Plappert and Jerry Tworek and Jacob Hilton and Reiichiro Nakano and Christopher Hesse and John Schulman},
      year={2021},
      eprint={2110.14168},
      archivePrefix={arXiv},
      primaryClass={cs.LG},
      url={https://arxiv.org/abs/2110.14168}, 
}

@inproceedings{kadlcik-etal-2023-calc,
    title = "Calc-{X} and Calcformers: Empowering Arithmetical Chain-of-Thought through Interaction with Symbolic Systems",
    author = "Kadl{\v{c}}{\'i}k, Marek  and
      {\v{S}}tef{\'a}nik, Michal  and
      Sotolar, Ondrej  and
      Martinek, Vlastimil",
    editor = "Bouamor, Houda  and
      Pino, Juan  and
      Bali, Kalika",
    booktitle = "Proceedings of the 2023 Conference on Empirical Methods in Natural Language Processing",
    month = dec,
    year = "2023",
    address = "Singapore",
    publisher = "Association for Computational Linguistics",
    url = "https://aclanthology.org/2023.emnlp-main.742/",
    doi = "10.18653/v1/2023.emnlp-main.742",
    pages = "12101--12108"
}

@inproceedings{patel-etal-2021-nlp,
    title = "Are {NLP} Models really able to Solve Simple Math Word Problems?",
    author = "Patel, Arkil  and
      Bhattamishra, Satwik  and
      Goyal, Navin",
    editor = "Toutanova, Kristina  and
      Rumshisky, Anna  and
      Zettlemoyer, Luke  and
      Hakkani-Tur, Dilek  and
      Beltagy, Iz  and
      Bethard, Steven  and
      Cotterell, Ryan  and
      Chakraborty, Tanmoy  and
      Zhou, Yichao",
    booktitle = "Proceedings of the 2021 Conference of the North American Chapter of the Association for Computational Linguistics: Human Language Technologies",
    month = jun,
    year = "2021",
    address = "Online",
    publisher = "Association for Computational Linguistics",
    url = "https://aclanthology.org/2021.naacl-main.168/",
    doi = "10.18653/v1/2021.naacl-main.168",
    pages = "2080--2094"
}

@inproceedings{NEURIPS2024_3d5aa9a7,
 author = {Toshniwal, Shubham and Moshkov, Ivan and Narenthiran, Sean and Gitman, Daria and Jia, Fei and Gitman, Igor},
 booktitle = {Advances in Neural Information Processing Systems},
 doi = {10.52202/079017-1096},
 editor = {A. Globerson and L. Mackey and D. Belgrave and A. Fan and U. Paquet and J. Tomczak and C. Zhang},
 pages = {34737--34774},
 publisher = {Curran Associates, Inc.},
 title = {OpenMathInstruct-1: A 1.8 Million Math Instruction Tuning Dataset},
 url = {https://proceedings.neurips.cc/paper_files/paper/2024/file/3d5aa9a7ce28cdc710fbd044fd3610f3-Paper-Datasets_and_Benchmarks_Track.pdf},
 volume = {37},
 year = {2024}
}

@inproceedings{
toshniwal2025openmathinstruct,
title={OpenMathInstruct-2: Accelerating {AI} for Math with Massive Open-Source Instruction Data},
author={Shubham Toshniwal and Wei Du and Ivan Moshkov and Branislav Kisacanin and Alexan Ayrapetyan and Igor Gitman},
booktitle={The Thirteenth International Conference on Learning Representations},
year={2025},
url={https://openreview.net/forum?id=mTCbq2QssD}
}

@misc{numina_math_datasets,
  author = {Jia LI and Edward Beeching and Lewis Tunstall and Ben Lipkin and Roman Soletskyi and Shengyi Costa Huang and Kashif Rasul and Longhui Yu and Albert Jiang and Ziju Shen and Zihan Qin and Bin Dong and Li Zhou and Yann Fleureau and Guillaume Lample and Stanislas Polu},
  title = {NuminaMath},
  year = {2024},
  publisher = {Numina},
  journal = {Hugging Face repository},
  howpublished = {\url{[https://huggingface.co/AI-MO/NuminaMath-CoT](https://github.com/project-numina/aimo-progress-prize/blob/main/report/numina_dataset.pdf)}}
}

@misc{zeng2024skyworkmathdatascalinglaws,
      title={Skywork-Math: Data Scaling Laws for Mathematical Reasoning in Large Language Models -- The Story Goes On}, 
      author={Liang Zeng and Liangjun Zhong and Liang Zhao and Tianwen Wei and Liu Yang and Jujie He and Cheng Cheng and Rui Hu and Yang Liu and Shuicheng Yan and Han Fang and Yahui Zhou},
      year={2024},
      eprint={2407.08348},
      archivePrefix={arXiv},
      primaryClass={cs.AI},
      url={https://arxiv.org/abs/2407.08348}, 
}

@inproceedings{
gao2025omnimath,
title={Omni-{MATH}: A Universal Olympiad Level Mathematic Benchmark for Large Language Models},
author={Bofei Gao and Feifan Song and Zhe Yang and Zefan Cai and Yibo Miao and Qingxiu Dong and Lei Li and Chenghao Ma and Liang Chen and Runxin Xu and Zhengyang Tang and Benyou Wang and Daoguang Zan and Shanghaoran Quan and Ge Zhang and Lei Sha and Yichang Zhang and Xuancheng Ren and Tianyu Liu and Baobao Chang},
booktitle={The Thirteenth International Conference on Learning Representations},
year={2025},
url={https://openreview.net/forum?id=yaqPf0KAlN}
}

@inproceedings{
ye2025limo,
title={{LIMO}: Less is More for Reasoning},
author={Yixin Ye and Zhen Huang and Yang Xiao and Ethan Chern and Shijie Xia and Pengfei Liu},
booktitle={Second Conference on Language Modeling},
year={2025},
url={https://openreview.net/forum?id=T2TZ0RY4Zk}
}

@misc{moshkov2025aimo2winningsolutionbuilding,
      title={AIMO-2 Winning Solution: Building State-of-the-Art Mathematical Reasoning Models with OpenMathReasoning dataset}, 
      author={Ivan Moshkov and Darragh Hanley and Ivan Sorokin and Shubham Toshniwal and Christof Henkel and Benedikt Schifferer and Wei Du and Igor Gitman},
      year={2025},
      eprint={2504.16891},
      archivePrefix={arXiv},
      primaryClass={cs.AI},
      url={https://arxiv.org/abs/2504.16891}, 
}

@inproceedings{jin-etal-2024-impact,
    title = "The Impact of Reasoning Step Length on Large Language Models",
    author = "Jin, Mingyu  and
      Yu, Qinkai  and
      Shu, Dong  and
      Zhao, Haiyan  and
      Hua, Wenyue  and
      Meng, Yanda  and
      Zhang, Yongfeng  and
      Du, Mengnan",
    editor = "Ku, Lun-Wei  and
      Martins, Andre  and
      Srikumar, Vivek",
    booktitle = "Findings of the Association for Computational Linguistics: ACL 2024",
    month = aug,
    year = "2024",
    address = "Bangkok, Thailand",
    publisher = "Association for Computational Linguistics",
    url = "https://aclanthology.org/2024.findings-acl.108/",
    doi = "10.18653/v1/2024.findings-acl.108",
    pages = "1830--1842"
}

@inproceedings{
sprague2025to,
title={To CoT or not to CoT? Chain-of-thought helps mainly on math and symbolic reasoning},
author={Zayne Rea Sprague and Fangcong Yin and Juan Diego Rodriguez and Dongwei Jiang and Manya Wadhwa and Prasann Singhal and Xinyu Zhao and Xi Ye and Kyle Mahowald and Greg Durrett},
booktitle={The Thirteenth International Conference on Learning Representations},
year={2025},
url={https://openreview.net/forum?id=w6nlcS8Kkn}
}

@inproceedings{tutek-etal-2025-measuring,
    title = "Measuring Chain of Thought Faithfulness by Unlearning Reasoning Steps",
    author = "Tutek, Martin  and
      Hashemi Chaleshtori, Fateme  and
      Marasovic, Ana  and
      Belinkov, Yonatan",
    editor = "Christodoulopoulos, Christos  and
      Chakraborty, Tanmoy  and
      Rose, Carolyn  and
      Peng, Violet",
    booktitle = "Proceedings of the 2025 Conference on Empirical Methods in Natural Language Processing",
    month = nov,
    year = "2025",
    address = "Suzhou, China",
    publisher = "Association for Computational Linguistics",
    url = "https://aclanthology.org/2025.emnlp-main.504/",
    doi = "10.18653/v1/2025.emnlp-main.504",
    pages = "9946--9971",
    ISBN = "979-8-89176-332-6"
}

@inproceedings{
hu2022lora,
title={Lo{RA}: Low-Rank Adaptation of Large Language Models},
author={Edward J Hu and yelong shen and Phillip Wallis and Zeyuan Allen-Zhu and Yuanzhi Li and Shean Wang and Lu Wang and Weizhu Chen},
booktitle={International Conference on Learning Representations},
year={2022},
url={https://openreview.net/forum?id=nZeVKeeFYf9}
}

@inproceedings{10.5555/3692070.3694581,
author = {Zhao, Hao and Andriushchenko, Maksym and Croce, Francesco and Flammarion, Nicolas},
title = {Long is more for alignment: a simple but tough-to-beat baseline for instruction fine-tuning},
year = {2024},
publisher = {JMLR.org},
booktitle = {Proceedings of the 41st International Conference on Machine Learning},
articleno = {2511},
numpages = {30},
location = {Vienna, Austria},
series = {ICML'24}
}

@inproceedings{koksal-etal-2024-longform,
    title = "{L}ong{F}orm: Effective Instruction Tuning with Reverse Instructions",
    author = {K{\"o}ksal, Abdullatif  and
      Schick, Timo  and
      Korhonen, Anna  and
      Schuetze, Hinrich},
    editor = "Al-Onaizan, Yaser  and
      Bansal, Mohit  and
      Chen, Yun-Nung",
    booktitle = "Findings of the Association for Computational Linguistics: EMNLP 2024",
    month = nov,
    year = "2024",
    address = "Miami, Florida, USA",
    publisher = "Association for Computational Linguistics",
    url = "https://aclanthology.org/2024.findings-emnlp.414/",
    doi = "10.18653/v1/2024.findings-emnlp.414",
    pages = "7056--7078"
}

@misc{deepseekai2025deepseekr1incentivizingreasoningcapability,
      title={DeepSeek-R1: Incentivizing Reasoning Capability in LLMs via Reinforcement Learning}, 
      author={DeepSeek-AI},
      year={2025},
      eprint={2501.12948},
      archivePrefix={arXiv},
      primaryClass={cs.CL},
      url={https://arxiv.org/abs/2501.12948}, 
}

@inproceedings{
ye2025data,
title={Data Mixing Laws: Optimizing Data Mixtures by Predicting Language Modeling Performance},
author={Jiasheng Ye and Peiju Liu and Tianxiang Sun and Jun Zhan and Yunhua Zhou and Xipeng Qiu},
booktitle={The Thirteenth International Conference on Learning Representations},
year={2025},
url={https://openreview.net/forum?id=jjCB27TMK3}
}

@inproceedings{
liu2025regmix,
title={RegMix: Data Mixture as Regression for Language Model Pre-training},
author={Qian Liu and Xiaosen Zheng and Niklas Muennighoff and Guangtao Zeng and Longxu Dou and Tianyu Pang and Jing Jiang and Min Lin},
booktitle={The Thirteenth International Conference on Learning Representations},
year={2025},
url={https://openreview.net/forum?id=5BjQOUXq7i}
}

@misc{wang2025demystifyinghybridthinkingllms,
      title={Demystifying Hybrid Thinking: Can LLMs Truly Switch Between Think and No-Think?}, 
      author={Shouren Wang and Wang Yang and Xianxuan Long and Qifan Wang and Vipin Chaudhary and Xiaotian Han},
      year={2025},
      eprint={2510.12680},
      archivePrefix={arXiv},
      primaryClass={cs.LG},
      url={https://arxiv.org/abs/2510.12680}, 
}

@inproceedings{
li2025data,
title={Data Mixing Optimization for Supervised Fine-Tuning of Large Language Models},
author={Yuan Li and Zhengzhong Liu and Eric P. Xing},
booktitle={Forty-second International Conference on Machine Learning},
year={2025},
url={https://openreview.net/forum?id=19kqoNoc2N}
}

@misc{gemmateam2025gemma3technicalreport,
      title={Gemma 3 Technical Report}, 
      author={Gemma Team},
      year={2025},
      eprint={2503.19786},
      archivePrefix={arXiv},
      primaryClass={cs.CL},
      url={https://arxiv.org/abs/2503.19786}, 
}

@inproceedings{zhang-etal-2025-adaptthink,
    title = "{A}dapt{T}hink: Reasoning Models Can Learn When to Think",
    author = "Zhang, Jiajie  and
      Lin, Nianyi  and
      Hou, Lei  and
      Feng, Ling  and
      Li, Juanzi",
    editor = "Christodoulopoulos, Christos  and
      Chakraborty, Tanmoy  and
      Rose, Carolyn  and
      Peng, Violet",
    booktitle = "Proceedings of the 2025 Conference on Empirical Methods in Natural Language Processing",
    month = nov,
    year = "2025",
    address = "Suzhou, China",
    publisher = "Association for Computational Linguistics",
    url = "https://aclanthology.org/2025.emnlp-main.184/",
    doi = "10.18653/v1/2025.emnlp-main.184",
    pages = "3716--3730",
    ISBN = "979-8-89176-332-6"
}

@inproceedings{
arora2025training,
title={Training Language Models to Reason Efficiently},
author={Daman Arora and Andrea Zanette},
booktitle={The Thirty-ninth Annual Conference on Neural Information Processing Systems},
year={2025},
url={https://openreview.net/forum?id=AiZxn84Wdo}
}

@article{team2025kimi,
  title={Kimi k1. 5: Scaling reinforcement learning with llms},
  author={Team Kimi and Du, Angang and Gao, Bofei and Xing, Bowei and Jiang, Changjiu and Chen, Cheng and Li, Cheng and Xiao, Chenjun and Du, Chenzhuang and Liao, Chonghua and others},
  journal={arXiv preprint arXiv:2501.12599},
  year={2025}
}

@inproceedings{shen-etal-2025-dast,
    title = "{DAST}: Difficulty-Adaptive Slow-Thinking for Large Reasoning Models",
    author = "Shen, Yi  and
      Zhang, Jian  and
      Huang, Jieyun  and
      Shi, Shuming  and
      Zhang, Wenjing  and
      Yan, Jiangze  and
      Wang, Ning  and
      Wang, Kai  and
      Liu, Zhaoxiang  and
      Lian, Shiguo",
    editor = "Potdar, Saloni  and
      Rojas-Barahona, Lina  and
      Montella, Sebastien",
    booktitle = "Proceedings of the 2025 Conference on Empirical Methods in Natural Language Processing: Industry Track",
    month = nov,
    year = "2025",
    address = "Suzhou (China)",
    publisher = "Association for Computational Linguistics",
    url = "https://aclanthology.org/2025.emnlp-industry.160/",
    doi = "10.18653/v1/2025.emnlp-industry.160",
    pages = "2322--2331",
    ISBN = "979-8-89176-333-3"
}

@article{wu2025unlocking,
  title={Unlocking efficient long-to-short llm reasoning with model merging},
  author={Wu, Han and Yao, Yuxuan and Liu, Shuqi and Liu, Zehua and Fu, Xiaojin and Han, Xiongwei and Li, Xing and Zhen, Hui-Ling and Zhong, Tao and Yuan, Mingxuan},
  journal={arXiv preprint arXiv:2503.20641},
  year={2025}
}

\appendix

% \section{Example Appendix}
% \label{sec:appendix}

% This is an appendix.

% \subsection{Prompts for Training}

\section{Experimental Settings}\label{App:training setting}

\paragraph{Baseline.}
To study whether the non-thinking and thinking modes affect each other's performance, we adopt a TMF training setting with a data ratio of T:NT = 1:1 and a training schedule of NT-T as the baseline.
To analyze the effect of training schedules on Thinking Mode Fusion, we use the performance of the base model without any fine-tuning as the baseline.
To study the effect of data ratios on Thinking Mode Fusion, we again adopt TMF training with a data ratio of T:NT = 1:1 and a training schedule of T-NT as the baseline.

\paragraph{Overview of Training settings.}
All experiments use continual supervised fine-tuning (SFT) on Qwen3-4B with Transformers and TRL.\footnote{https://huggingface.co/docs/transformers}\footnote{https://huggingface.co/docs/trl}
To isolate TMF effects while maintaining efficiency, we apply LoRA~\citep{hu2022lora} to the main projection modules (\{q\_proj, k\_proj, v\_proj, o\_proj, gate\_proj, up\_proj, down\_proj\}) with $r{=}16$, $\alpha{=}64$, dropout $0.0$, and no bias.
Optimization uses AdamW with a cosine learning rate schedule, a learning rate of $2\times10^{-5}$, weight decay $0.001$, and 5 warmup steps.
All runs use a maximum sequence length of $5{,}120$ tokens and train for one epoch; results are averaged over three runs. %(see Appendix~\ref{App:training setting} for all details).

\subsection{Base Model and Training Framework}
We conduct all experiments starting from \textbf{Qwen3-4B}, an open-weight model that natively supports a unified chat template with optional \texttt{<think>} blocks. All fine-tuning runs are implemented using Transformers and TRL with SFTTrainer. To isolate the effects of Thinking Mode Fusion (TMF) strategies, we perform continual supervised fine-tuning (SFT) with a lightweight parameter-efficient setup via LoRA \citep{hu2022lora}.

Specifically, we apply LoRA to all major projection modules in the transformer blocks:
\texttt{\{q\_proj, k\_proj, v\_proj, o\_proj, gate\_proj, up\_proj, down\_proj\}}.
Unless otherwise stated, LoRA uses rank $r{=}16$, $\alpha{=}64$, dropout $0.0$, and no bias adaptation. Training is performed in \textbf{bfloat16} with \texttt{device\_map=auto}. We enable TF32 for CUDA matmul and cuDNN when available to improve throughput.

\subsection{Training Data}
To instantiate the two modes in TMF, we use one dataset for long-form reasoning supervision (thinking mode) and one dataset for short-solution supervision (non-thinking mode).

\paragraph{Thinking mode data.}
We use \textbf{OpenMathReasoning}, split \texttt{cot}) as our thinking mode dataset. Each sample includes a math \texttt{problem} and a long chain-of-thought style \texttt{generated\_solution}. We randomly split the dataset into 80\% training and 20\% test using a fixed seed.

\paragraph{Non-thinking mode data.}
We use \textbf{GSM8K}, configuration \texttt{main}) as our non-thinking dataset. Each sample contains a \texttt{question} and an \texttt{answer} with a canonical ``\texttt{\#\#\#\#}'' delimiter. We use the official train/test split.

\paragraph{Training subset size.}
To ensure controlled comparisons across different fusion ratios, we cap the thinking mode training set to a fixed size $N_T=\texttt{standard\_index}$ (default 1500). For each run, we sample the non-thinking set size $N_{NT}$ according to a target ratio $\rho = N_T / N_{NT}$ (denoted as \texttt{reasoning\_chat\_percentage} in code), i.e.,
\begin{equation}
N_{NT} = \left\lfloor N_T \cdot \frac{1}{\rho} \right\rfloor .
\end{equation}
Non-thinking samples are randomly sub-sampled with a fixed random seed for reproducibility.

\subsection{Example of Data}\label{App:Example of Data}

Thinking mode data:

\begin{quote}
Suppose that ( x, y in mathbb{R} ) satisfy ( |x-1| leq 2 ) and ( |2y+1| leq 5 ). Let ( M ) and ( m ) be the maximum and minimum of ( xy - x + 3y + 3 ), respectively. Find the value of ( M + m ).
\end{quote}

Non-thinking mode data:

\begin{quote}
Natalia sold clips to 48 of her friends in April, and then she sold half as many clips in May. How many clips did Natalia sell altogether in April and May?
\end{quote}

\subsection{Prompting and Mode Formatting}\label{App:Prompting and Mode Formattin}
We unify both datasets under a chat-style supervision format using the model’s native chat template. For thinking mode data, the user instruction explicitly requests step-by-step reasoning and enforces the final answer format:
\begin{quote}
\small \texttt{Please reason step by step, and put your final answer within \textbackslash boxed\{\}. 
Problem: 
Suppose that ( x, y in mathbb{R} ) satisfy ( |x-1| leq 2 ) and ( |2y+1| leq 5 ). Let ( M ) and ( m ) be the maximum and minimum of ( xy - x + 3y + 3 ), respectively. Find the value of ( M + m ).}
\end{quote}

\begin{examplebox}
\detokenize{
<think>Okay, so I need to find the maximum and minimum values of the expression xy - x + 3y + 3 given the constraints |x - 1|} $\le$ \detokenize{2 and |2y + 1|} $\le$ \detokenize{5. Then, once I have those, I need to add them together to get M + m. Let me try to break this down step by step. First, let's understand the constraints. The first one is |x - 1|} $\le$ \detokenize{2. This means that x is within 2 units of 1. So, solving that inequality, we subtract 1 and take the absolute value into account: |x - 1|} $\le$ \detokenize{2 implies -2} $\le$ \detokenize{x - 1} $\le$ \detokenize{2. ...
</think> Given the constraints...Therefore, the final answer is:boxed{-6}.
}
\end{examplebox}

% \begin{examplebox}
% % <think>Okay, so I need to find the maximum and minimum values of the expression xy - x + 3y + 3 given the constraints |x - 1| ≤ 2 and |2y + 1| ≤ 5. Then, once I have those, I need to add them together to get M + m. Let me try to break this down step by step. First, let's understand the constraints. The first one is |x - 1| ≤ 2. This means that x is within 2 units of 1. So, solving that inequality, we subtract 1 and take the absolute value into account: |x - 1| ≤ 2 implies -2 ≤ x - 1 ≤ 2. ...</think> Given the constraints...Therefore, the final answer is:boxed{-6}.\\\
% \end{examplebox}

For non-thinking mode data, the instruction requests a direct solution with the same final answer constraint:
\begin{quote}
\small \texttt{Solve the problem, and put your final answer within \textbackslash boxed\{\}.
Problem: 
Natalia sold clips to 48 of her friends in April, and then she sold half as many clips in May. How many clips did Natalia sell altogether in April and May?}
\end{quote}

\begin{examplebox}
<think></think> Natalia sold 48/2 = <<48/2=24>>24 clips in May.Natalia sold 48+24 = <<48+24=72>>72 clips altogether in April and May.**Final Answer** boxed\{72\}.
\end{examplebox}

\begin{figure}[t]
\centering
\subfigure[Non-thinking.]{\includegraphics[width=0.9\linewidth]{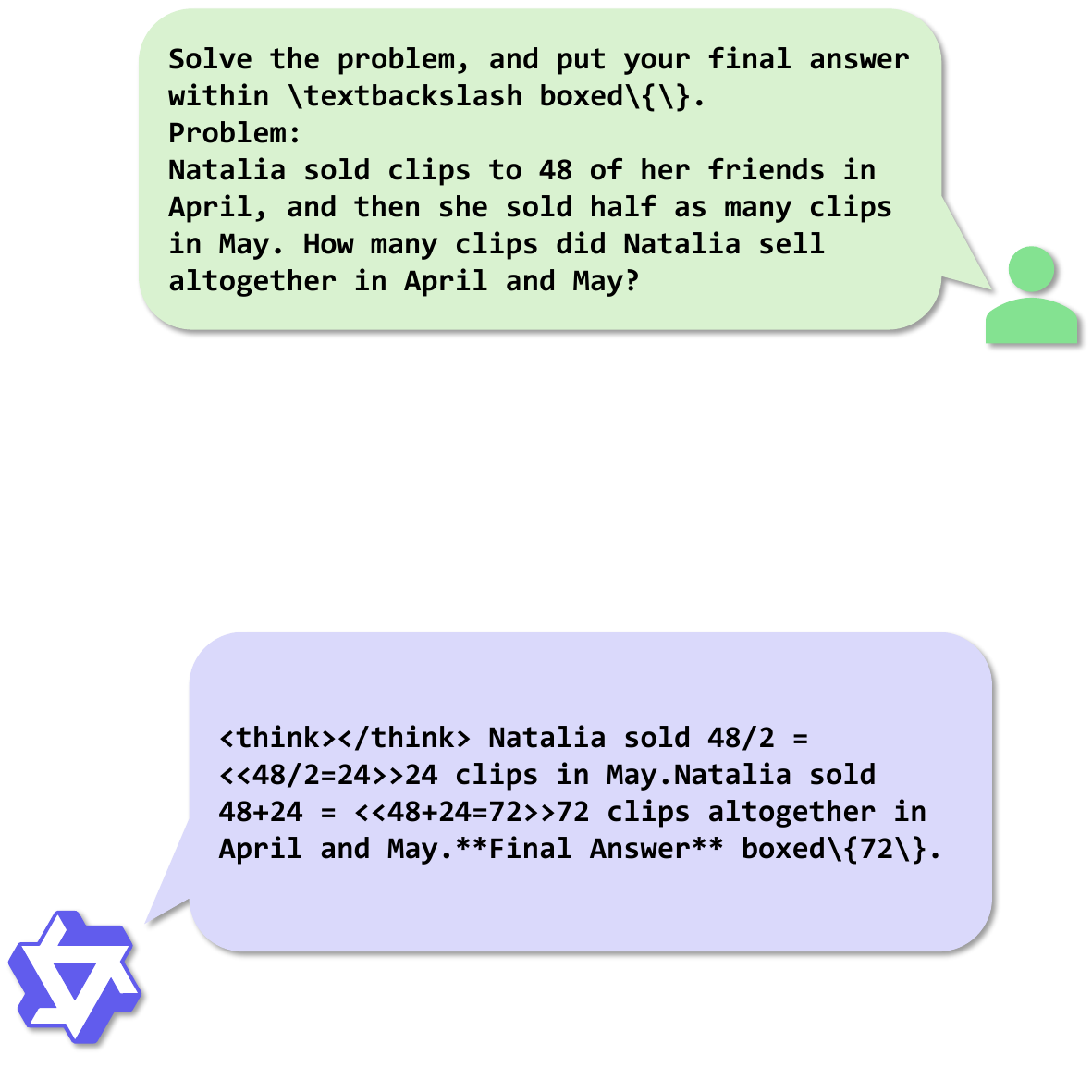}}
\subfigure[Thinking.]{\includegraphics[width=0.9\linewidth]{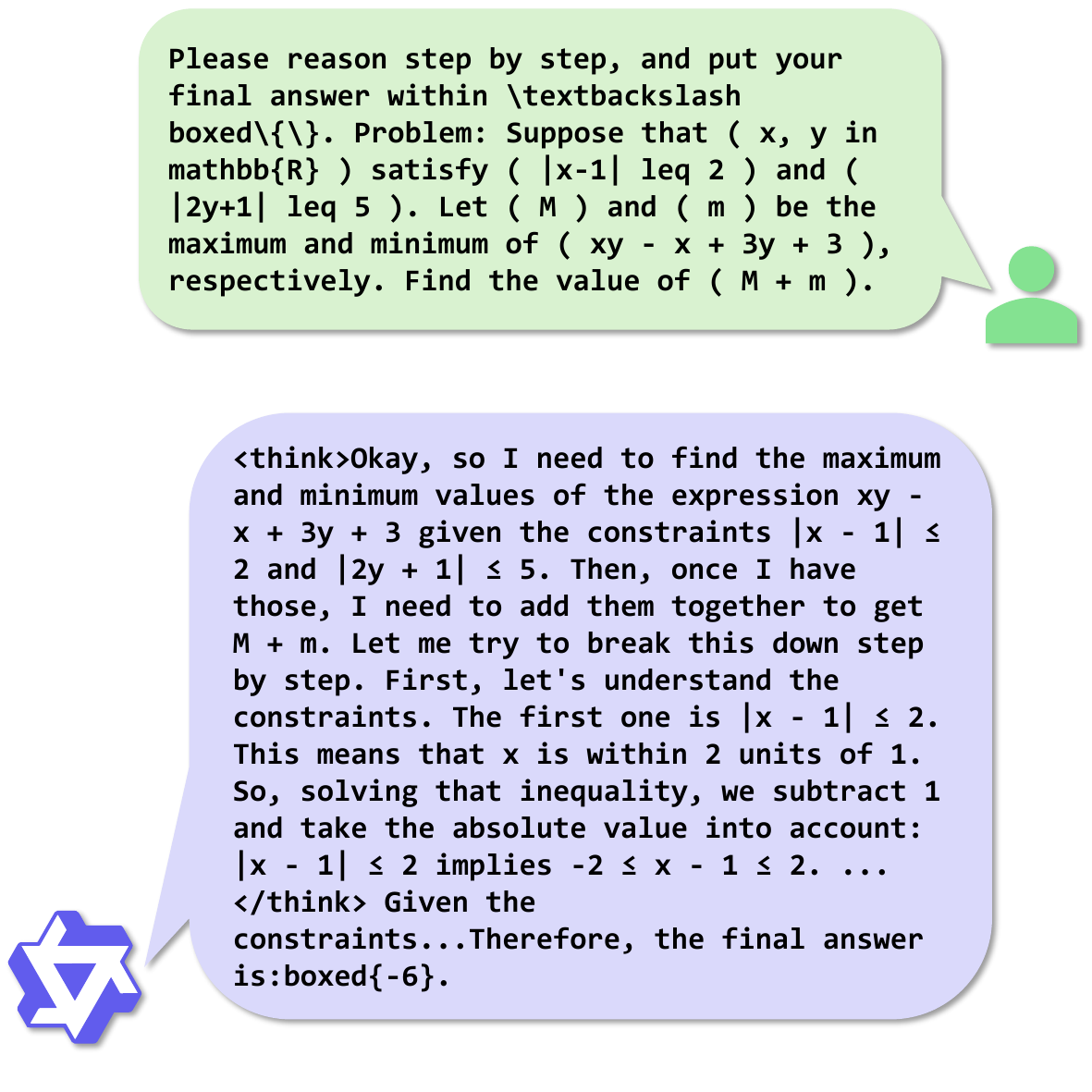}}
\caption{An overview of the experimental design for analyzing Thinking Mode Fusion under different training orders and data ratios.}
\label{fig:fig6}
\end{figure}

To align with the TMF interface, we wrap non-thinking solutions with an \texttt{<think>} block that is intentionally left empty, followed by a short solution and a boxed final answer. This design enforces a consistent output schema across modes while discouraging long reasoning in the non-thinking mode.

\subsection{TMF Training Strategies: Order and Mixing}
We study three training strategies that control the interaction between thinking and non-thinking supervision:

\begin{itemize}
    \item \textbf{T-NT}: sequential training where thinking mode samples are placed before non-thinking samples in the SFT stream.
    \item \textbf{NT-T}: sequential training where non-thinking samples are placed before thinking samples.
    \item \textbf{Mix}: an interleaving strategy that alternates the two modes according to the target ratio $\rho$. When $\rho \le 1$ (i.e., thinking is less frequent), we interleave \texttt{1 thinking : $n$ non-thinking} with $n \approx 1/\rho$. When $\rho > 1$, we interleave \texttt{$n$ thinking : 1 non-thinking} with $n \approx \rho$.
\end{itemize}

All strategies share the same total number of training examples determined by $(N_T, N_{NT})$ under each ratio.

\subsection{Optimization and Hyperparameters}
We fine-tune the model using AdamW with cosine learning-rate scheduling. Unless otherwise stated, hyperparameters are fixed across all runs:
\begin{itemize}
    \item Learning rate: $2\times 10^{-5}$
    \item Weight decay: $0.001$
    \item Warmup steps: 5
    \item Epochs: 1
    \item Per-device batch size: 1
    \item Gradient accumulation: 6 (effective batch size 6)
    \item Maximum sequence length: 5120 tokens
\end{itemize}

We enable gradient checkpointing during training and disable key-value caching to reduce memory usage.

\subsection{Evaluation Protocol}
We evaluate on both tasks to quantify interference and transfer between modes: GSM8K test for non-thinking capability and OpenMathReasoning-mini test for thinking capability. For efficiency and controlled comparisons, we sub-sample each test set to $\lfloor 0.25 \cdot \texttt{standard\_index} \rfloor$ examples.

\paragraph{Generation settings.}
We use greedy decoding (\texttt{do\_sample=False}) with specific maximum generation budgets:
\texttt{max\_new\_tokens=1024} for GSM8K and \texttt{max\_new\_tokens=5120} for OpenMathReasoning. The final answer is extracted via pattern matching on the last occurrence of \texttt{\textbackslash boxed\{...\}}.

\paragraph{Throughput-aware batching and long-tail control.}
To stabilize evaluation cost across different prompt lengths, we estimate prompt token statistics on a random sample (default 1024 prompts) and set the evaluation prefill truncation threshold to the \textbf{95th percentile} token length (capped at 8192). Inputs longer than the percentile threshold are evaluated with batch size 1 to avoid padding inefficiency; extremely long prompts beyond the hard cap are filtered by default. We additionally report decoding throughput in tokens/second.

\paragraph{Metric.}
We report exact-match accuracy between the extracted boxed prediction and the gold answer. All evaluations are conducted at the end of each epoch via a custom callback and logged to Weights \& Biases, together with token statistics and throughput.

\subsection{Run Management}
All experiments are executed on a single-node, single-GPU (NVIDIA H100) cluster setting using SLURM array jobs. We sweep over fusion strategies and data ratios, and repeat each configuration multiple times with separate run identifiers. Each run logs configurations and evaluation metrics to Weights \& Biases for aggregation and analysis.

% =================================================

\end{document}